\documentclass[runningheads]{llncs}

\usepackage[mobile]{eccv}
\usepackage{wrapfig}
\usepackage{multirow}

\usepackage{eccvabbrv}

\usepackage{graphicx}
\usepackage{booktabs}

\usepackage[accsupp]{axessibility}  

\usepackage[pagebackref,breaklinks,colorlinks]{hyperref}

\usepackage{orcidlink}

\begin{document}

\title{Make-Your-3D: Fast and Consistent Subject-Driven 3D Content Generation}

\author{
Fangfu Liu \and
Hanyang Wang \and
Weiliang Chen \and 
Haowen Sun \and
Yueqi Duan\thanks{\small{Corresponding author.}}
}
\authorrunning{Liu et al.}
\institute{
Tsinghua University
}





\maketitle

\vspace{-0.5em}
\begin{center}
    \centering
    \includegraphics[width=\linewidth]{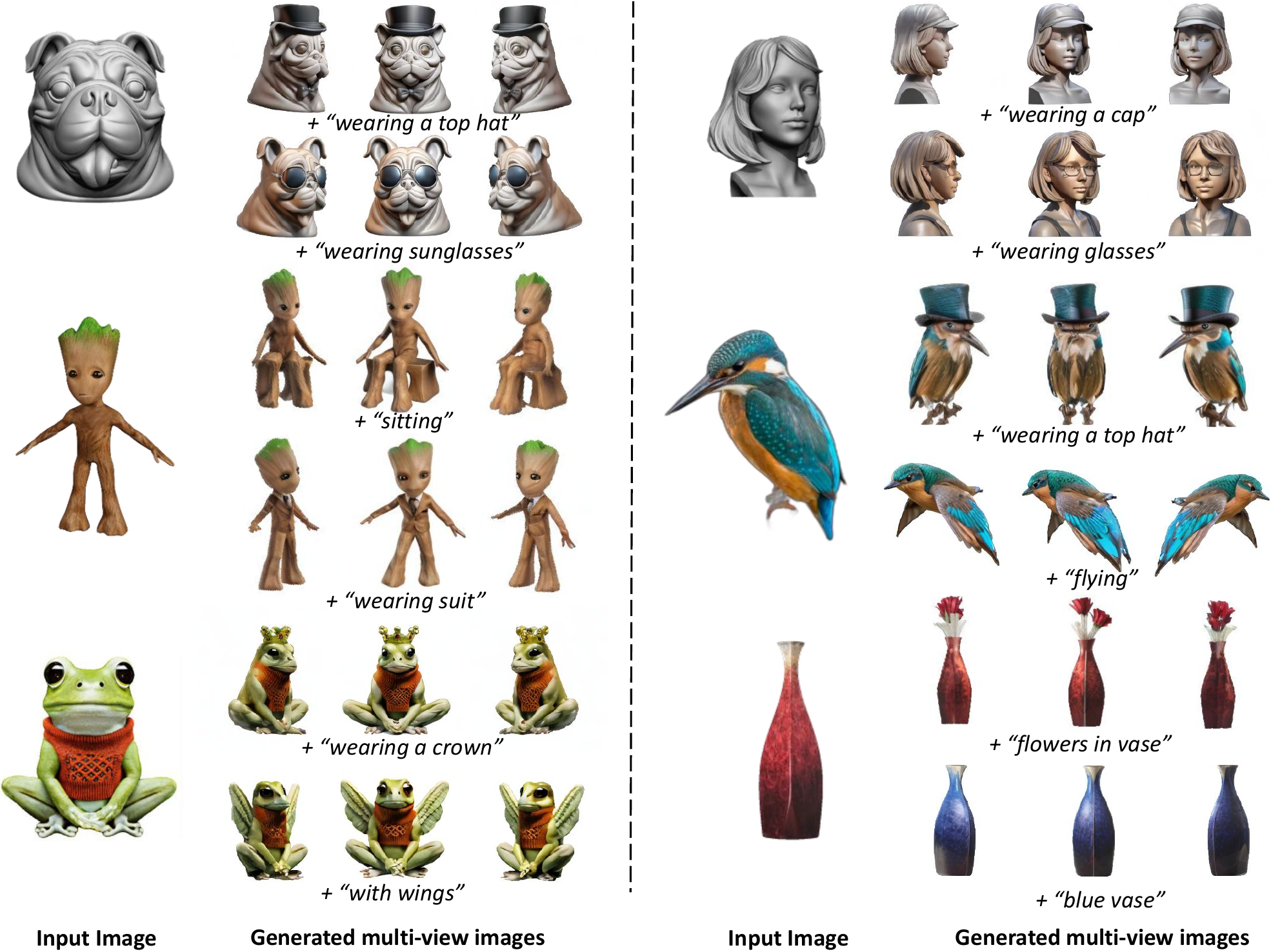}%
    \captionof{figure}
    {
    \textbf{Make-Your-3D} can personalize 3D contents from only a \textbf{single image} of a subject with text-driven modifications within only \textbf{5 minutes}. 
    }
    \label{fig:teaser}
\end{center}%

\begin{abstract}
   Recent years have witnessed the strong power of 3D generation models, which offer a new level of creative flexibility by allowing users to guide the 3D content generation process through a single image or natural language. However, it remains challenging for existing 3D generation methods to create subject-driven 3D content across diverse prompts. In this paper, we introduce a novel 3D customization method, dubbed \textbf{Make-Your-3D} that can personalize high-fidelity and consistent 3D content from only a single image of a subject with text description within 5 minutes. Our key insight is to harmonize the distributions of a multi-view diffusion model and an identity-specific 2D generative model, aligning them with the distribution of the desired 3D subject. Specifically, we design a co-evolution framework to reduce the variance of distributions, where each model undergoes a process of learning from the other through identity-aware optimization and subject-prior optimization, respectively. Extensive experiments demonstrate that our method can produce high-quality, consistent, and subject-specific 3D content with text-driven modifications that are unseen in subject image. Project page: \url{https://liuff19.github.io/Make-Your-3D/}.

\end{abstract}
\section{Introduction}
Subject-driven customization has emerged as a prominent aspect within the field of generative models~\cite{ruiz2023dreambooth, yang2023diffusion-survey, zeng2023avatarbooth, wu2023tune-a-video, raj2023dreambooth3d}, providing a wide range of multimedia applications~\cite{zhang2023text-diffusion-survey, wu2023aigc-survey}. It aims to synthesize the individual subject in diverse contexts and styles while retaining high fidelity to subject-specific identities. Despite the great progress of personalization in text-to-image (T2I)~\cite{ruiz2023dreambooth, ruiz2023hyperdreambooth, ye2023ip-adapter, chen2024subject-apprenticeship, avrahami2023chosen-one} and text-to-video (T2V)~\cite{wu2023tune-a-video, xing2023make-your-video, wei2023dreamvideo, ren2024customizeavideo, ma2024magic-me} models, the exploration of customized 3D generation remains relatively limited.

Driven by the advancements in neural 3D representations~\cite{kerbl20233d-gaussian, mildenhall2021nerf-eccv}, extensive datasets~\cite{deitke2023objaverse, deitke2024objaverse-xl, chang2015shapenet}, and the diffusion-based generative models~\cite{ho2020ddpm, rombach2022stable-diffusion}, recent works~\cite{poole2022dreamfusion, wang2023prolificdreamer, long2023wonder3d, tang2023dreamgaussian, shi2023mvdream, lin2023magic3d} have demonstrated high-quality automated 3D generation from text or image prompt. Although text or image prompts allow for some degree of control over the generated 3D asset, it remains challenging to produce high-fidelity and subject-specific 3D content with text-driven modifications such as novel colors, poses, or attributes that are unseen in any of the input subject images. Enabling the creation of subject-specific 3D assets through flexible text controls would greatly simplify the workflow for artists and streamline 3D acquisition processes. One notable attempt for subject-driven 3D generation is DreamBooth3D~\cite{raj2023dreambooth3d}, which combines a personalizing model (DreamBooth~\cite{ruiz2023dreambooth}) and a text-to-3D model (DreamFusion~\cite{poole2022dreamfusion}) with two-stage tuning for DreamBooth to optimize NeRF~\cite{mildenhall2021nerf-eccv} representation. However, it has two inherent limitations: (a) time-consuming optimization of NeRF representation with the two fine-tuning stages in DreamBooth and (b) requirements of multiple subject-specific images as input, which significantly limits the range of applications. 

In this paper, we propose \textbf{Make-Your-3D}, a novel co-evolution framework for fast and consistent subject-driven 3D content generation. Specifically, given only a single casual subject image, we can generate subject-specific 3D assets that align with text-driven modifications as contextualization within 5 minutes, which is \textbf{36$\times$ faster }than DreamBooth3D~\cite{raj2023dreambooth3d}. As shown in Fig.~\ref{fig:teaser}, we personalize 3D content with geometric consistency and strong appearance identity preservation from given subjects while also respecting the variations (\eg, sitting or wearing suit) provided by input text prompts. 

For \text{Make-Your-3D}, we draw inspiration from recent advancements in personalized models~\cite{ye2023ip-adapter} and multi-view diffusion models~\cite{long2023wonder3d}. Despite the customization capability of personalized models and the 3D consistency of multi-view diffusion models, there remains a domain gap between the targeted subject and these two models, particularly when the subject is unseen in the training data~\cite{ruiz2023dreambooth, shi2023mvdream}. Therefore, the key idea of our method is to harmonize the distribution of the identity-specific 2D generative model and multi-view diffusion model, aligning them with the distribution of the desired 3D subject. Specifically, we design a co-evolution framework to reduce the variance of distributions, where each model undergoes a process of learning from the other through identity-aware optimization and subject-prior optimization, respectively. Given a casual image captured from a subject, we first lift it to a 3D space through a multi-view diffusion model and capture its multi-views to optimize the 2D personalized model with an identity-enhancement process, which imposes enhanced identity-aware information into the 2D model. Next, we apply the original 2D personalized model to multi-views of the subject with modified text description and obtain more diverse images of the subject. Then we optimize the multi-view diffusion model with such various subject images in the subject-prior optimization process, infusing the subject-specific prior into the multi-view diffusion model. Finally, we cascade the two optimized models to process the single input image of the targeted subject and generate consistent subject-driven 3D results.

We conduct extensive experiments on the dataset used in DreamBooth3D~\cite{ruiz2023dreambooth} and open-vocabulary wild images captured from a subject with different styles. We also conduct a user study to evaluate the subject and prompt fidelity in our synthesized 3D results. The experiments validate that our method is capable of producing vivid and high-fidelity 3D assets with strong adherence to the given subject while highly respecting the contextualization in the input text prompts. Compared to DreamBooth3D~\cite{raj2023dreambooth3d}, our method not only surpasses in terms of quality, resolution, and consistency, but also shows a remarkable 36$\times$ speed improvement for efficiency. Unlike DreamBooth3D~\cite{raj2023dreambooth3d}, our approach takes only a single wild image as input, eliminating the need for 3-6 carefully selected images of the same subject. Fig.~\ref{fig:teaser} shows sample results of Make-Your-3D on different subjects and contextualizations, indicating that our co-evolution framework promotes the capability of generating subject-specific 3D assets for both models.

\section{Related Works}
\label{sec:formatting}
\textbf{Text-to-3D Generation.}
With exciting breakthroughs emerging in the image and video generation~\cite{zhang2023texttoimage, xing2023survey, 10419041, ho2020denoising, reed2016generative}, there has been growing interest in 3D content generation~\cite{liu2024comprehensive, li2024advances}, particularly in text-to-3D generation~\cite{li2023generative}. One approach is to utilize extensive data~\cite{chang2015shapenet, deitke2023objaverse, deitke2024objaverse-xl} to train 3D generative models~\cite{chen2023singlestage, gupta20233dgen, kim2023neuralfieldldm, lorraine2023att3d, luo2021diffusion, nichol2022pointe, shue20223d, zhang20233dshape2vecset, zhao2023michelangelo}, akin to Text-to-Image (T2I) generation. However, due to constraints by the scale and quality of paired text-3D data, these methods are often limited to specific object categories and may exhibit a perceived lack of realism. Another line of text-to-3D is pioneered by DreamFusion~\cite{poole2022dreamfusion}, which employed a Score Distillation Sampling (SDS) loss to optimize a parametric 3D representation guided by the pre-trained 2D diffusion model~\cite{saharia2022photorealistic-imagen}. Following works concentrate on improving 3D consistency, novel view quality, and generation speed through strategies such as incorporating 3D priors~\cite{tang2023mvdiffusion, shi2023mvdream, liu2023sherpa3d}, crafting a tailored optimization strategy~\cite{sun2023dreamcraft3d, wang2023prolificdreamer,lin2023magic3d}, and selecting more expressive and efficient representations~\cite{chen2023fantasia3d, tang2023dreamgaussian}. However, only text is not informative enough to express complex scenes or concepts, which can be a hindrance to 3D content creation. 
Moreover, it is often unattainable to create contextually diverse 3D assets that precisely align with the user-desired objects.
\newline 

\noindent\textbf{Image-to-3D  Generation.}
Given the rich information embedded in images, numerous studies~\cite{deng2022nerdi,xu2023neurallift360,melaskyriazi2023realfusion,tang2023makeit3d,tang2023dreamgaussian,zhang2023repaint123,long2023wonder3d,woo2023harmonyview} have explored to generate 3D content from a single image. Early attempts integrated the input image into the optimization pipeline by creating loss based on predicted depth~\cite{deng2022nerdi,xu2023neurallift360} or object masks~\cite{melaskyriazi2023realfusion}, comparing them with the rendered image. Magic123~\cite{qian2023magic123} designs a two-stage coarse-to-fine framework for high-quality image-to-3D generation, employing textual inversion to ensure the generation of object-preserving geometry and textures. DreamGaussian~\cite{tang2023dreamgaussian} and \text{Repaint123}~\cite{zhang2023repaint123} leverage a more efficient Gaussian splatting representation~\cite{kerbl20233d-gaussian}, significantly improving the optimization speed. Wonder3D~\cite{long2023wonder3d} uses a 2D diffusion model to generate multi-view normal maps with color images and applies a geometry-aware normal fusion algorithm for direct surface extraction. However, despite the capability of these approaches to generate 3D content from a single image, their excessive reliance on images to maintain 3D consistency results in a lack of diversity in the generated 3D content, sometimes resembling more of a 3D reconstruction task. While HarmonyView~\cite{woo2023harmonyview} introduces the concept of harmonizing consistency and diversity, the displayed results often fall short of delivering satisfactory diversity, let alone achieving subject-driven customization. Different from their methods, our work is dedicated to reconstructing the concept of the provided object rather than the input image, thereby preserving the generated diversity.
\newline

\noindent\textbf{Subject-Driven Content Creation.}
An increasing number of works~\cite{gal2022image, kumari2023multiconcept, ruiz2023dreambooth, jiang2023videobooth, raj2023dreambooth3d, huang2024customizeit3d} are focusing on subject-driven generation, enabling users to personalize the generated content for specific subjects and concepts. Dreambooth~\cite{ruiz2023dreambooth} finetunes the 2D diffusion model and expands the model's language-vision dictionary with rare tokens using multiple images, achieving personalized text-to-image generation. IP-Adapter~\cite{ye2023ip-adapter} realizes controllable generation by incorporating image prompt in the text-to-image models via the design of a lightweight decoupled cross-attention mechanism. VideoBooth~\cite{jiang2023videobooth} injects image prompts into the text-to-video model (T2V) in a coarse-to-fine manner, achieving customized content generation for videos. Despite remarkable success in personalizing T2I and T2V models, they do not generate 3D assets or afford any 3D control.
The first attempt in 3D subject-driven generation is DreamBooth3D~\cite{raj2023dreambooth3d}, which proposes a simple pipeline utilizing DreamBooth~\cite{ruiz2023dreambooth} for 3D subject-driven generation. However, its generation is constrained by DreamBooth~\cite{ruiz2023dreambooth}, requiring heavy fine-tuning stages with multiple subject images over 3 hours, which limits the range of applications. In contrast, our method can achieve fast subject-driven 3D content generation from only a single image of a subject within 5 minutes.

\section{Method}
\begin{figure*}[!t]
    \centering
    \includegraphics[width=\linewidth]{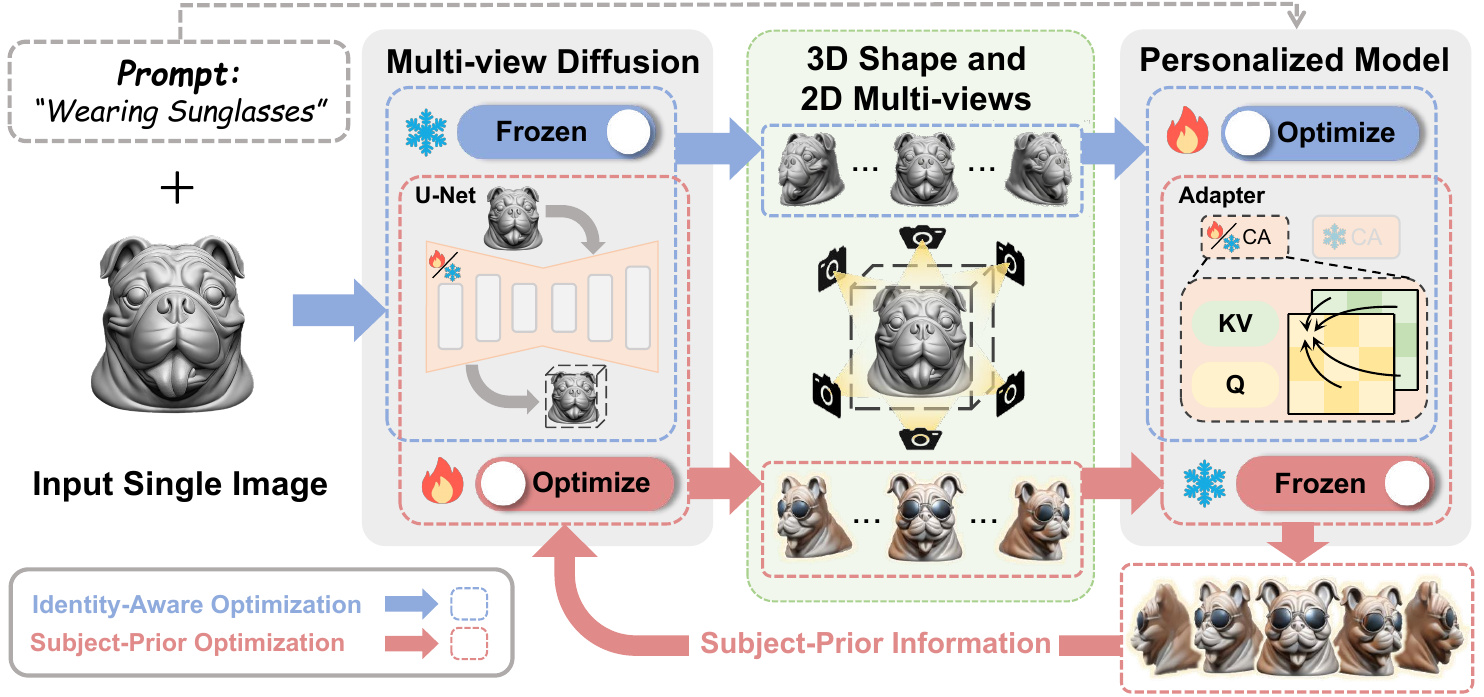}
    \caption{\textbf{The overall framework of our proposed Make-Your-3D.} Our framework includes identity-aware optimization of 2D personalized model and subject-prior optimization of multi-view diffusion model to approximate subject distribution. The identity-aware optimization (Sec.~\ref{subsec: identity-aware opt}) lifts input image to 3D space through a frozen multi-view diffusion model and optimizes the 2D personalized model via multi-views. The subject-prior optimization (Sec.~\ref{subsec: subject-prior opt}) adopts diverse images from frozen personalized model to infuse the subject-specific prior into the multi-view diffusion model.}
    \label{fig:pipeline}
\end{figure*}

In this section, we introduce our framework, \textit{i.e.}, Make-Your-3D, for fast and consistent subject-driven 3D content generation. Our goal is to align output 3D assets with the distribution of the desired subject. To this end, we first review the scheme of diffusion model (Sec.~\ref{subsec:preliminary}), which is the basis of our pre-trained multi-view and personalized models. Then we analyze the distribution variance and optimization target of our method (Sec.~\ref{subsec:distribution}). We further introduce our co-evolution framework, including identity-aware optimization (Sec.~\ref{subsec: identity-aware opt}) and subject-prior optimization (Sec.~\ref{subsec: subject-prior opt}). Finally, we present our mesh extraction process (Sec.~\ref{subsec:mesh}). An overview of our framework is depicted in Fig.~\ref{fig:pipeline}.

\begin{figure*}[!t]
    \centering
    \includegraphics[width=\linewidth]{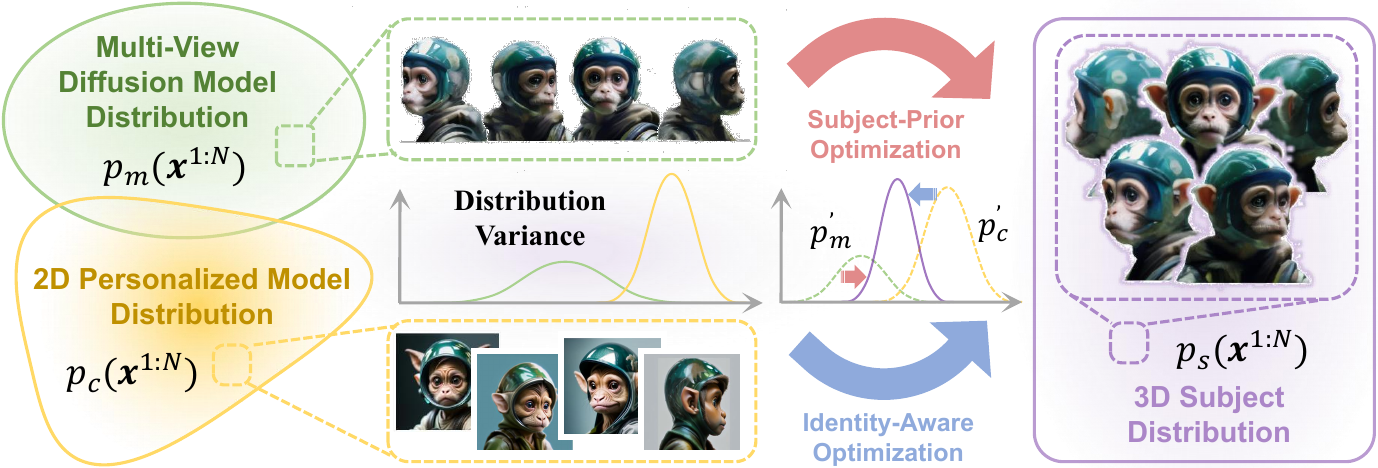}
    \caption{\textbf{Distribution variance between the wild subject and pre-trained models.} Taking a \textit{monkey} image and text prompt \textit{``with elf ears''} as input, the pre-trained 2D personalized model and multi-view diffusion model generate images out of the distribution of desired ones, \textit{i.e.}, the specific monkey with elf ears. To solve the problem, we carefully design a co-evolution framework including subject-prior and identity-aware optimization to harmonize the distributions and achieves desired 3D assets.}
    \label{fig:distribution}
\end{figure*}

\subsection{Preliminaries}
\label{subsec:preliminary}
\noindent \textbf{Diffusion Models}~\cite{sohl2015deep-diffusion, ho2020ddpm} are probabilistic generative models that comprise two processes: (a) a forward process that gradually adds Gaussian noise to the data following a T steps Markov chain and (b) a denoising process that generates samples from the Gaussian distribution. 
Let $\boldsymbol{x}_0 \sim q(\boldsymbol{x}_0)$ represent the real data under the additional condition $\boldsymbol{c}$, and let $t \in [0, T]$ denote the time step of the diffusion process. The training objective of the diffusion model $\boldsymbol{\epsilon}_\theta$, which predicts noise, is calculated as the following variational bound:
\begin{equation}
    \mathcal{L}_{\textit{diff}} = \mathbb{E}_{\boldsymbol{x}_0, \boldsymbol{\epsilon} \sim \mathcal{N}(\mathbf{0}, \mathbf{I}), \boldsymbol{c}, t}\left\|\boldsymbol{\epsilon}-\boldsymbol{\epsilon}_\theta\left(\boldsymbol{x}_t, \boldsymbol{c}, t\right)\right\|^2, 
    \label{eq: diffusion-loss}
\end{equation}
where $\boldsymbol{x}_t = \alpha_t \boldsymbol{x}_0 + \sigma_t \boldsymbol{\epsilon}$ is the noisy data at step $t$, and $\alpha_t, \sigma_t$ are fixed sequence of the noise schedule. For the conditional diffusion models, classifier-free guidance~\cite{ho2022classifier-free-guidance} is often employed as a prevailing method. In the sampling stage, the prediction noise is computed as a combination of conditional model $\boldsymbol{\epsilon}_\theta\left(\boldsymbol{x}_t, \boldsymbol{c}, t\right)$ and unconditional model $\boldsymbol{\epsilon}_\theta\left(\boldsymbol{x}_t, \boldsymbol{c}\right)$:
\begin{equation}
    \hat{\boldsymbol{\epsilon}}_\theta\left(\boldsymbol{x}_t, \boldsymbol{c}, t\right)=w \boldsymbol{\epsilon}_\theta\left(\boldsymbol{x}_t, \boldsymbol{c}, t\right)+(1-w) \boldsymbol{\epsilon}_\theta\left(\boldsymbol{x}_t, t\right),
\end{equation}
where $w$ is the guidance scale to adjust the alignment with condition $\boldsymbol{c}$. In our study, we utilize the open-source 2D personalized model~\cite{ye2023ip-adapter} and the multi-view diffusion model~\cite{long2023wonder3d}, which are both built upon the Stable-Diffusion model~\cite{rombach2022stable-diffusion}, to achieve fast and consistent subject-driven 3D generation. 

\subsection{The Distribution of 3D Subject}
\label{subsec:distribution}
Powered by recent advances in personalizing text-to-image models~\cite{ruiz2023dreambooth, ye2023ip-adapter, ruiz2023hyperdreambooth} and image-to-3D models~\cite{wang2023imagedream, qian2023magic123, long2023wonder3d}, an intuitive idea to achieve customized 3D generation is to naively combine these methods. However, as shown in Fig.~\ref{fig:distribution}, it fails to yield satisfactory subject-specific 3D assets due to distribution variance between the wild subject and the pre-trained models~\cite{ruiz2023dreambooth, ye2023ip-adapter, du2024stable-is-unstable}. To explore how to approximate the subject domain, we first propose that the distribution of the 3D subject denoted as $q_s(\boldsymbol{z})$, can be modeled as a joint distribution as:
\begin{equation}
    q_s(\boldsymbol{z}) = p_s(\boldsymbol{x}^{1:N} | \mathcal{I}, y)=p_s(\boldsymbol{x}^{1} | \mathcal{I}, y) \cdot p_s(\boldsymbol{x}^{2:N} | \mathcal{I}, y),
    \label{eq: q_s}
\end{equation}
where $\boldsymbol{x}^{1:N}$ are 2D multi-view color images observed from 3D subject conditioned on subject image $\mathcal{I}$ and text-driven modification $y$. As we have the pre-trained 2D customized model $p_c(\boldsymbol{x}^1|\mathcal{I}, y)$ and multi-view diffusion model $p_m(\boldsymbol{x}^{1:N}| \mathcal{I}) = \prod_{n=1}^{N} p_m(\boldsymbol{x}^{n} | \mathcal{I})$, our key insight involves optimizing both $p_c$ and $p_m$ to $p_c'$ and $p_m'$, which closely align with the distribution of $p_s$ respectively, \ie,  
\begin{equation}
    p_c'(\boldsymbol{x}^1|\mathcal{I}, y) \approx p_s(\boldsymbol{x}^{1} | \mathcal{I}, y) \, , \, p_m'(\boldsymbol{x}^{2:N}| \mathcal{I}) \approx p_s(\boldsymbol{x}^{2:N} | \mathcal{I}).
\end{equation}
Given that the condition $y$ is independent of $\boldsymbol{x}^{n}$ in the multi-view diffusion model, we can ultimately approximate $q_s(\boldsymbol{z})$ with the optimized models $p_c'$ and $p_m'$, \ie, $q_s(\boldsymbol{z}) \approx p_c' \cdot p_m'$. Finally, we can formulate the joint distribution as a Markov chain within the diffusion scheme (omit the symbol $y$ and $\mathcal{I}$ for simplicity):
\begin{equation}
    p_c'(\boldsymbol{x}_T^1)\, p_m'(\boldsymbol{x}_T^{2:N}) \prod_t p_m'(\boldsymbol{x}_{t-1}^1 | \boldsymbol{x}_{t}^1) \, p_m'(\boldsymbol{x}_{t-1}^{2:N}|\boldsymbol{x}_t^{2:N}),
\end{equation}
where $p_c'(\boldsymbol{x}_T^1)$ and $p_m'(\boldsymbol{x}_T^{2:N})$ are Gaussian noises. Our key problem is to characterize the distribution $p_c' \rightarrow p_s$ and $p_m' \rightarrow p_s$ so that we can sample from this Markov chain to generate 3D assets in the subject distribution.
Inspired by the derivation above, we carefully design a co-evolution framework, including identity-aware optimization for $p_c$ and subject-prior optimization for $p_m$.

\subsection{Identity-Aware Optimization}
\label{subsec: identity-aware opt}
To mimic the appearance of subjects from images and synthesize novel renditions of them in different contexts, DreamBooth~\cite{ruiz2023dreambooth} finetunes all the parameters of the T2I model with 3-6 captured subject images. However, such a tuning strategy that relies on several images, is inefficient and constrained to scenarios where only one image can serve as input. In contrast, we use only a single subject image as input and choose a more efficient adapter-based T2I model (\ie, IP-Adapter~\cite{ye2023ip-adapter}) as our 2D personalized model $p_c$. Despite the user-friendly appeal of using a single input image, the 2D personalized model~\cite{ye2023ip-adapter} suffers from distribution variance~\cite{du2024stable-is-unstable} between the subject and the training data, leading to outputs that do not resemble the subject as shown in Fig.~\ref{fig:ablation_study}, \ie, $p_c(\boldsymbol{x}^1|\mathcal{I}, y) \neq p_s(\boldsymbol{x}^{1}|\mathcal{I}, y)$. To approximate subject distribution and enhance awareness of identity, we first leverage a multi-view diffusion model~\cite{long2023wonder3d} $p_m$ to generate the multi-views $\boldsymbol{x}^{(1:N)}$ with view-direction aware prompt $y^{(1:N)}$ given the input subject image $\mathcal{I}$ and text-driven prompt $y$. Then we apply augmentations to $\boldsymbol{x}^{(1:N)}$ and process them through the pretrained CLIP image encoder~\cite{radford2021-CLIP} $\mathcal{F}$ and get $\mathcal{F}(\boldsymbol{x}^{(1:N)})$. Finally, in the adapter module of the 2D personalized model, we use $\mathcal{F}(\boldsymbol{x}^{(1:N)})$ and $y^{(1:N)}$ to optimize the parameters of image cross-attention layer while freezing the original UNet model and text cross-attention modules. We follow the similar training objective to obtain $p_c'$ in Eq.~\ref{eq: diffusion-loss} with the condition $\boldsymbol{c} = \{y, \mathcal{F}(\boldsymbol{x})\}$. 
The empirical analysis in Sec.~\ref{sec: experiments} demonstrates that our multi-view-based, identity-aware optimization effectively narrows the gap between $p_c$ and the subject domain $p_s$.

\subsection{Subject-Prior Optimization}
\label{subsec: subject-prior opt}
Powered by the 3D consistency of neural radiance fields, DreamBooth3D~\cite{raj2023dreambooth3d} distills the fine-tuned DreamBooth to generate 3D assets via score distillation sampling (SDS)~\cite{poole2022dreamfusion}. However, this framework suffers from low resolution and time-consuming optimization for per-sample training from scratch, as shown in Fig.~\ref{fig:compare_with_dreambooth3d},~\ref{fig:dreambooth3d_failure_cases}, limiting the practical usage. In this work, we optimize a more efficient multi-view diffusion framework based on Wonder3D~\cite{long2023wonder3d} to approximate $p_s(\boldsymbol{x}^{2:N} | \mathcal{I})$ while better achieving fast and high-fidelity personalized 3D generation. Given the multi-views $\boldsymbol{x}^{(1:N)}$ from subject image $\mathcal{I}$ as discussed in Sec.~\ref{subsec: identity-aware opt}, we process them through the original 2D personalized model with text-driven modification. Then we obtain diverse outputs $\tilde{\boldsymbol{x}}^{(1:N)}$ from multi-views, which coarsely adhere to the driven text and subject style with strong subject knowledge prior. In addition, we further exploit the subject geometry prior represented by normal maps $\tilde{\boldsymbol{n}}^{(1:N)}$ inferred from $\tilde{\boldsymbol{x}}^{(1:N)}$ by using the off-the-shelf single-view estimator~\cite{eftekhar2021omnidata-normal-estimator}. Finally, we optimize the cross-domain self-attention module in the UNet framework based on multi-view diffusion model~\cite{long2023wonder3d} to incorporate the subject-specific prior knowledge in the views of 3D distribution. Our objective function consists of two terms: (a) an image diffusion term for subject-prior enhancement and (b) a parameter preservation term for maintaining multi-view ability, which is computed as:
\begin{equation}
    \mathcal{L}_{\textit {prior }}=\mathbb{E}_{\boldsymbol{x}_0,\boldsymbol{n}_0, \boldsymbol{\epsilon}, \boldsymbol{c}_n, \boldsymbol{c}_i, t}\left\|\boldsymbol{\epsilon}-\boldsymbol{\epsilon}_\theta\left(\boldsymbol{x}_t, \boldsymbol{c}_{n}, \boldsymbol{c}_i, t\right)\right\|^2 + \lambda \frac{\left\|\theta-\theta_0 \right\|_1}{N_\theta},
    \label{eq: prior-loss}
\end{equation}
where $\boldsymbol{c}_i, \boldsymbol{c}_n$ are the condition of the subject image with corresponding normal maps, $\theta_0$ is the initial parameter of original multi-view diffusion, $N_\theta$ is the number of parameters, and $\lambda$ is a balancing parameter set to 1. Grounded by visualization studies in Sec.~\ref{sec: experiments}, our subject-prior optimization strategy can successfully impose subject prior knowledge into multi-view diffusion model, which aligns $p_m'(\cdot|\mathcal{I}) \rightarrow p_s(\cdot | \mathcal{I})$ by Eq.~\ref{eq: prior-loss} and yields more desired and consistent subject-driven 3D assets.

\subsection{Subject-Driven Mesh Extraction}
\label{subsec:mesh}
As done in previous co-evolution framework (\ie, identity-aware optimization and subject-prior optimization), we have aligned the 2D personalized model $p_c'$ and the multi-view diffusion model $p_m'$ with the subject distribution $p_s$. Given the subject image $\mathcal{I}$ and text modification $y$, we first cascade the two optimized models to process them and obtain subject-driven multiview color images $\hat{\boldsymbol{x}}^{(1:N)}$ with respect to Eq.~\ref{eq: q_s}. From $\hat{\boldsymbol{x}}^{(1:N)}$, we apply a recent U-Net based Gaussian model pretrained in LGM~\cite{tang2024lgm} to predict 3D Gaussians. Next, we train an efficient NeRF (\ie, Instant-NGP~\cite{muller2022instant-NGP}) by using the rendered images from 3D Gaussians, and then convert the NeRF to polygonal meshes~\cite{tang2023delicate-nerf-to-mesh}. More details can be found in our supplementary materials. With adequately optimized implementation in identity-aware optimization ($\sim 1$ min), subject-driven optimization ($\sim 3$ min), and mesh conversion ($\sim 1$ min), our framework can understand the visual subject in a reference image by approximating the subject distribution, and fast produce high-fidelity, consistent unseen personalized 3D content driven by text modification.

\section{Experiments}
\label{sec: experiments}
In this section, we conduct extensive experiments to evaluate our subject-driven 3D content generation framework Make-Your-3D, and show the comparison results against DreamBooth3D~\cite{raj2023dreambooth3d}. We first present our qualitative results in multi-views and comparisons with baselines~\cite{raj2023dreambooth3d, shi2023mvdream} in various applications (\eg, stylization, accessorization) (Sec.~\ref{subsec: qualitative compa}). Then we report the quantitative results with a user study (Sec.~\ref{subsec: quantitative result}). Finally, we carry out more open settings and ablation studies to further verify the efficacy of our framework design (Sec.~\ref{subsec: ablation}). Please refer to the supplementary materials for more visualizations, comparisons, and detailed analysis.

\subsection{Experiment Setup}
\noindent \textbf{Implementation Details.} In our framework implementation, we choose IP-Adapter~\cite{ye2023ip-adapter} as our 2D personalized model backbone and apply the learning rates of $1e-4$ with a $0.01$ weight decay to the image cross-attention layers with our multi-view based identity-aware optimization. On the other hand, we employ Wonder3D~\cite{long2023wonder3d} as our multi-view diffusion model and utilize diverse images generated by the original 2D personalized model to subject its U-Net module to subject-prior optimization, with a $5e-5$ learning rates and a $1e-2$ weight decay. Notably, for each subject image, it takes only 5 minutes to complete all optimization stages on a single NVIDIA RTX3090 (24GB) GPU, which is far more efficient than 3 hours tuning on 4 core TPUv4 used in DreamBooth3D~\cite{raj2023dreambooth3d}. We use a fixed 30 iterations to optimize the personalized model. For the multi-view diffusion model, we use around 100 iterations in subject-prior optimization across different objects. To reconstruct 3D geometry, our method is built on the instant-NGP~\cite{muller2022instant-NGP} based Gaussian reconstruction method~\cite{tang2023delicate-nerf-to-mesh}.
\newline 

\noindent \textbf{Baselines and Metrics.} We extensively compare our method with two baselines: DreamBooth3D~\cite{raj2023dreambooth3d} and an implementation for multi-view dreambooth in MVDream~\cite{shi2023mvdream}. Since the two baseline methods~\cite{raj2023dreambooth3d, shi2023mvdream} do not have released related code, their results are obtained by downloading from their project pages. For metrics, we mainly show our results with notable comparisons through visualization. Following~\cite{raj2023dreambooth3d, poole2022dreamfusion}, we evaluate our approach with the CLIP R-Precision metric in CLIP ViT-B/16, ViT-B/32, and ViT-L-14 models. We also conduct a user study to further demonstrate the subject-driven fidelity, prompt fidelity, consistency, and overall quality of our method.
\subsection{Qualitative Results}
\begin{figure*}[!t]
    \centering
    \includegraphics[width=\linewidth]{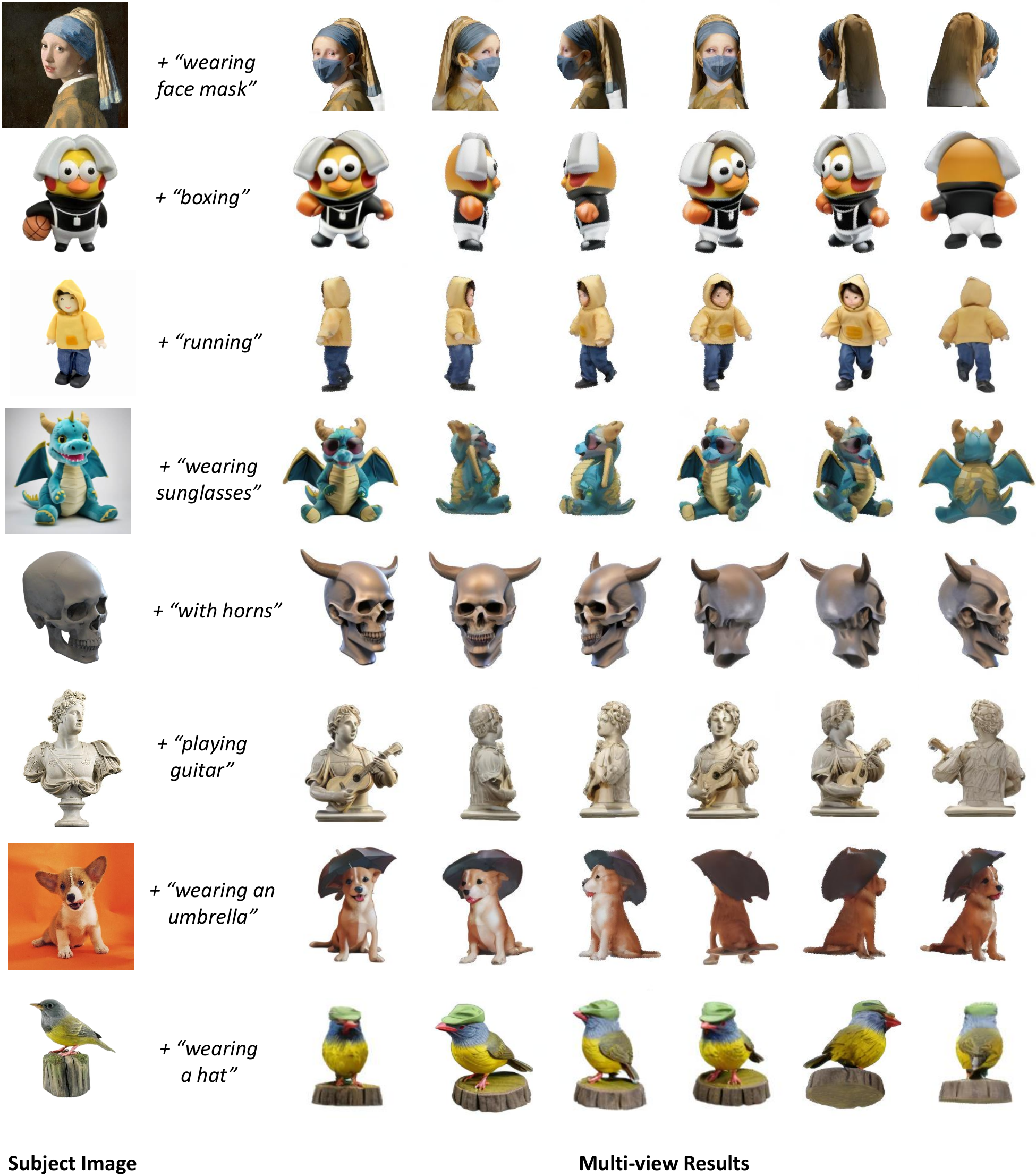}
    \caption{\textbf{Visual results of Make-Your-3D} on different subjects with customized text inputs. The multi-view results demonstrate that our method can generate 3D assets with high-fidelity, 3D consistency, subject preservation, and faithfulness to the text prompts.}
    \label{fig:visual_results}
    \vspace{-0.4cm}
\end{figure*}

\noindent \textbf{Visual Results of Make-Your-3D.} Fig.~\ref{fig:visual_results} shows sample visual results of our method across different subjects with customized text prompts. The results demonstrate high-fidelity and consistent 3D generation with Make-Your-3D for open-vocabulary wild input subject images, achieving faithful alignment respecting the context in the input text prompt.
\newline 

\begin{figure*}[!t]
    \centering
    \includegraphics[width=\linewidth]{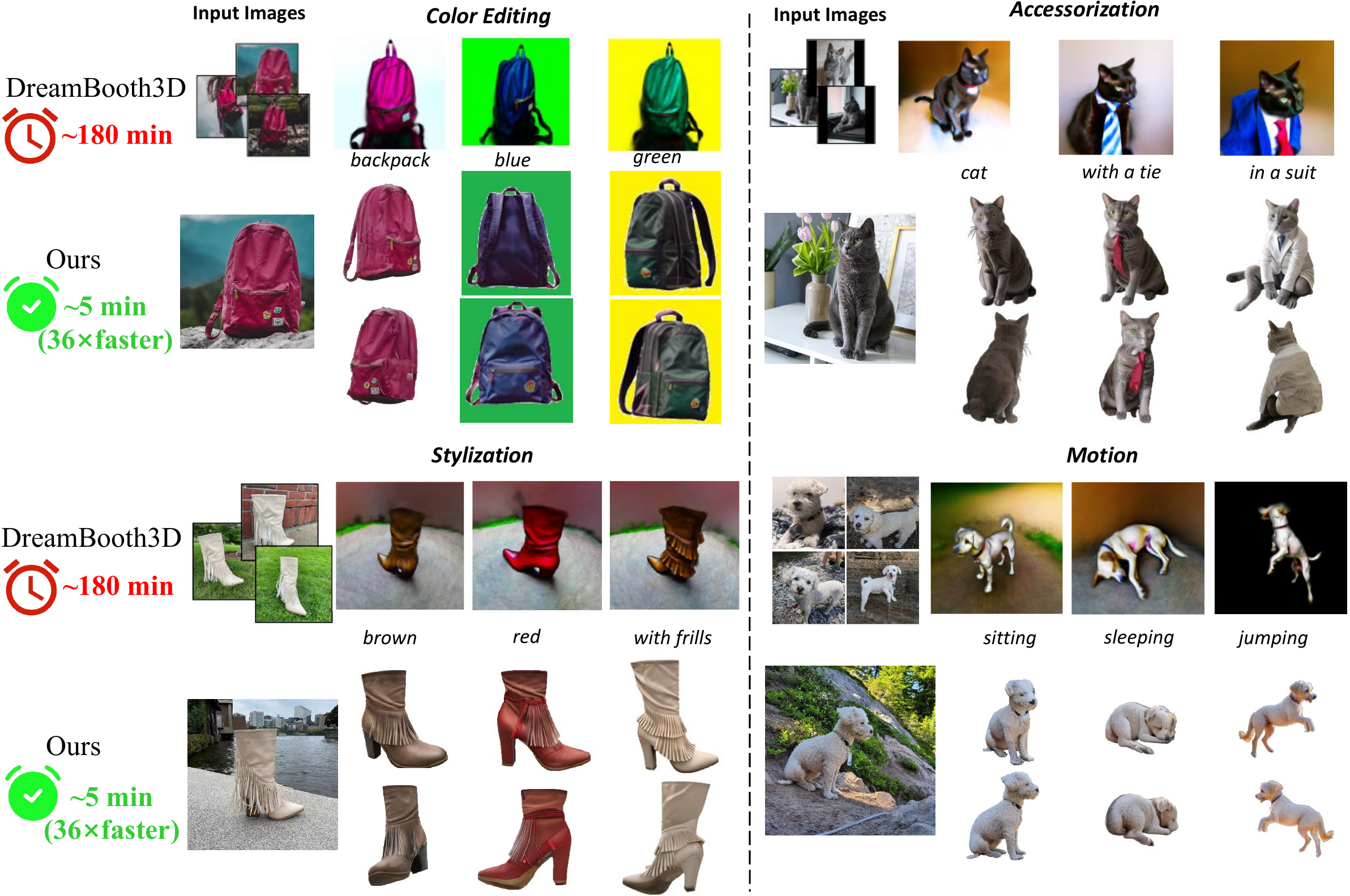}
    \caption{\textbf{The qualitative comparisons with DreamBooth3D}. We use the same text prompt and only one of the input images as in DreamBooth3D. Notice ours perform better on the object details with less input images.}
    \label{fig:compare_with_dreambooth3d}
\end{figure*}

\begin{figure*}[!t]
    \centering
    \includegraphics[width=\linewidth]{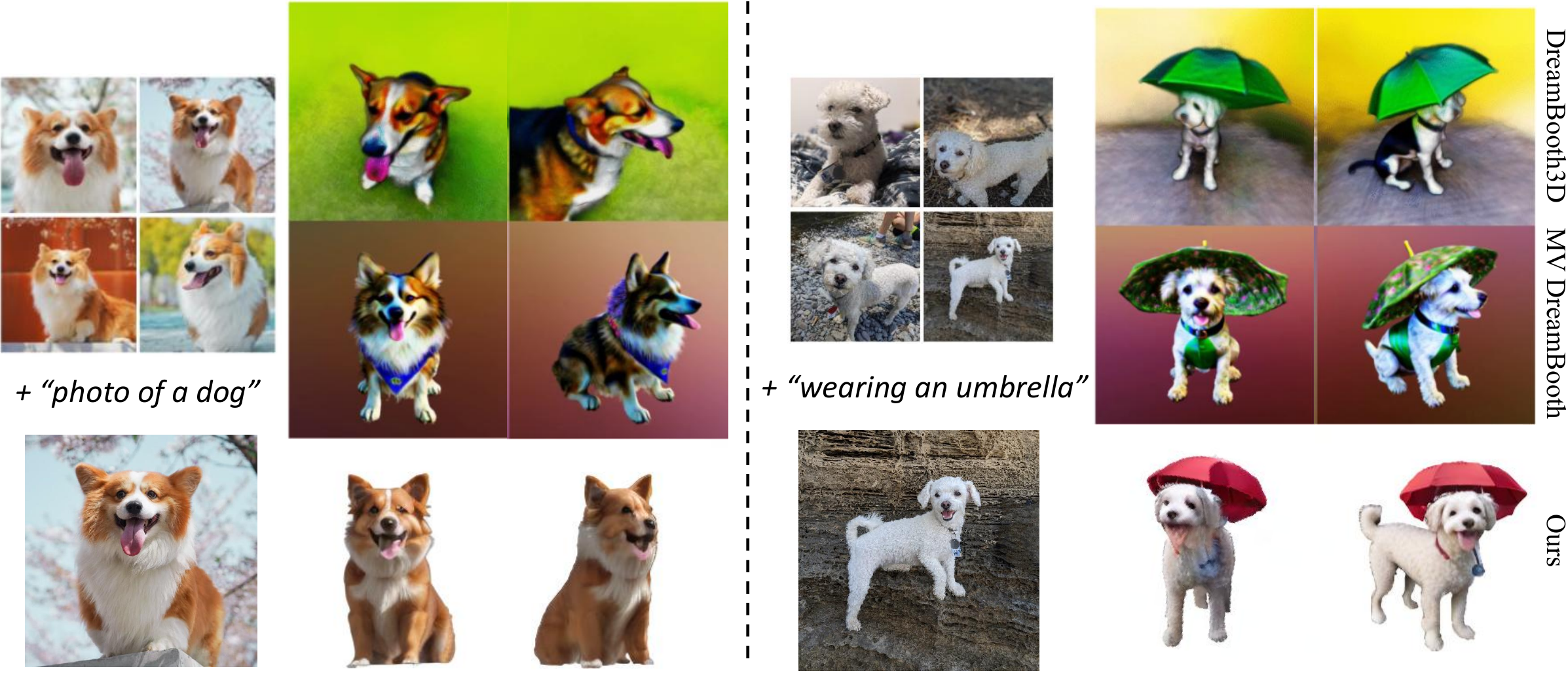}
    \caption{\textbf{The multi-view qualitative comparisons}. We are able to generate more realistic objects and achieve better subject preservation with multi-view consistency.}
    \label{fig:compare_with_mvdream}
\end{figure*}

\noindent \textbf{Qualitative Comparisons.} 
\label{subsec: qualitative compa}
We compare our method with DreamBooth3D~\cite{raj2023dreambooth3d} in various applications shown in Fig.~\ref{fig:compare_with_dreambooth3d}, including color editing, accessorization, stylization, and motion modification. We observe that DreamBooth3D only generates coarse results with multiple images as input over 3 hours. In contrast, our Make-Your-3D produces higher quality subject-driven 3D results with compelling object details from only \textbf{a single} subject image within only 5 minutes, which is \textbf{36$\times$} faster than DreamBooth3D. We also conduct comparisons with a recent multi-view DreamBooth implementation in MVDream~\cite{shi2023mvdream} in Fig.~\ref{fig:compare_with_mvdream}. The results further indicate that our method can not only achieve great 3D consistency but also better preserve subject identity without overfitting to data bias in terms of generated styles shown in mutli-view DreamBooth~\cite{shi2023mvdream}. 

\subsection{Quantitative Results}
\label{subsec: quantitative result}
\begin{wraptable}{r}{8.35cm}
    \centering
    \small
    \vspace{-11mm}
    \caption{\textbf{Quantitative comparisons} on rendered images with text prompts using different CLIP retrieval models.
    }
    \label{tab:precision_metrics}
    \vspace{1mm}
    \begin{tabular}{lccc}
        \toprule
        & ViT-B/16$\uparrow$ & ViT-B/32$\uparrow$ & ViT-L-14$\uparrow$ \\
        \midrule
        DreamBooth3D~\cite{raj2023dreambooth3d} & 0.783 & 0.710 & 0.797 \\
        MV DreamBooth~\cite{shi2023mvdream} & 0.805 & 0.735 & 0.813  \\
        Make-Your-3D (Ours) & \textbf{0.817} & \textbf{0.764} & \textbf{0.826}  \\
        \bottomrule
    \end{tabular}
    \vspace{-7mm}
\end{wraptable}
Table~\ref{tab:precision_metrics} shows the average CLIP R-Precision over 160 evenly spaces azimuth renders at a fixed elevation of 40 degrees, following the same setting in DreamBooth3D~\cite{raj2023dreambooth3d} for fairness. Results clearly demonstrate higher scores for Make-Your-3D, indicating better 3D consistency and text-prompt alignment of our results. For user study, we render 360-degree videos of subject-driven 3D models and show each volunteer with five samples of rendered video from a random method. They can rate in four aspects: 3D consistency, subject fidelity, prompt fidelity, and overall quality on a scale of 1-10, with higher
scores indicating better performance. We collect results from 30 volunteers shown in Table~\ref{tab:user_study}. We find our method is significantly preferred by users over these aspects. 

\begin{figure*}[!t]
    \centering
    \includegraphics[width=\linewidth]{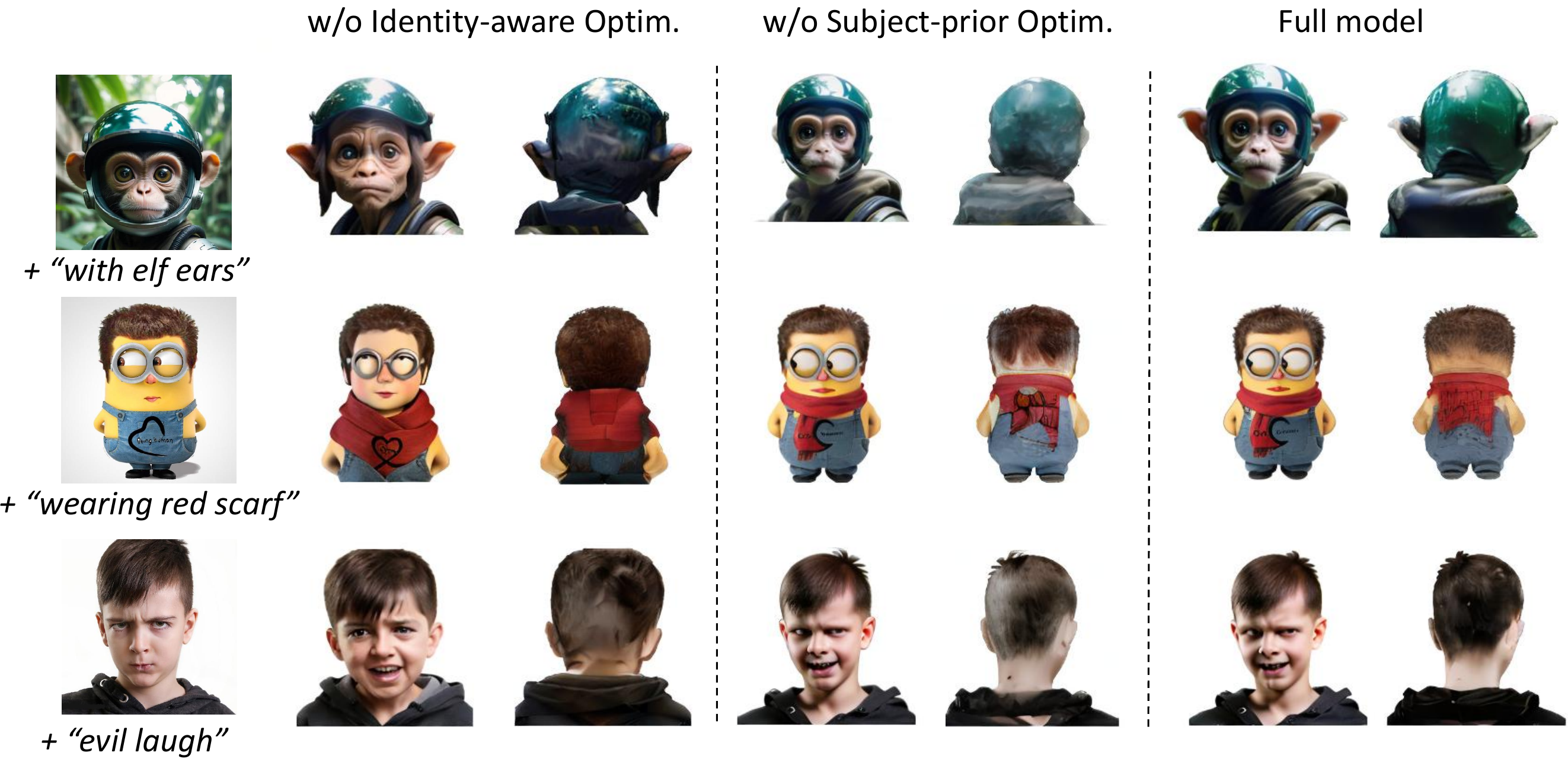}
    \caption{\textbf{Ablation study of our method}. We ablate the design choices of identity-aware optimization and subject-prior optimization.}
    \label{fig:ablation_study}
    \vspace{-2mm}
\end{figure*}

\subsection{Ablation Study and Discussion}
\label{subsec: ablation}
\begin{figure*}[!t]
    \centering
    \includegraphics[width=\linewidth]{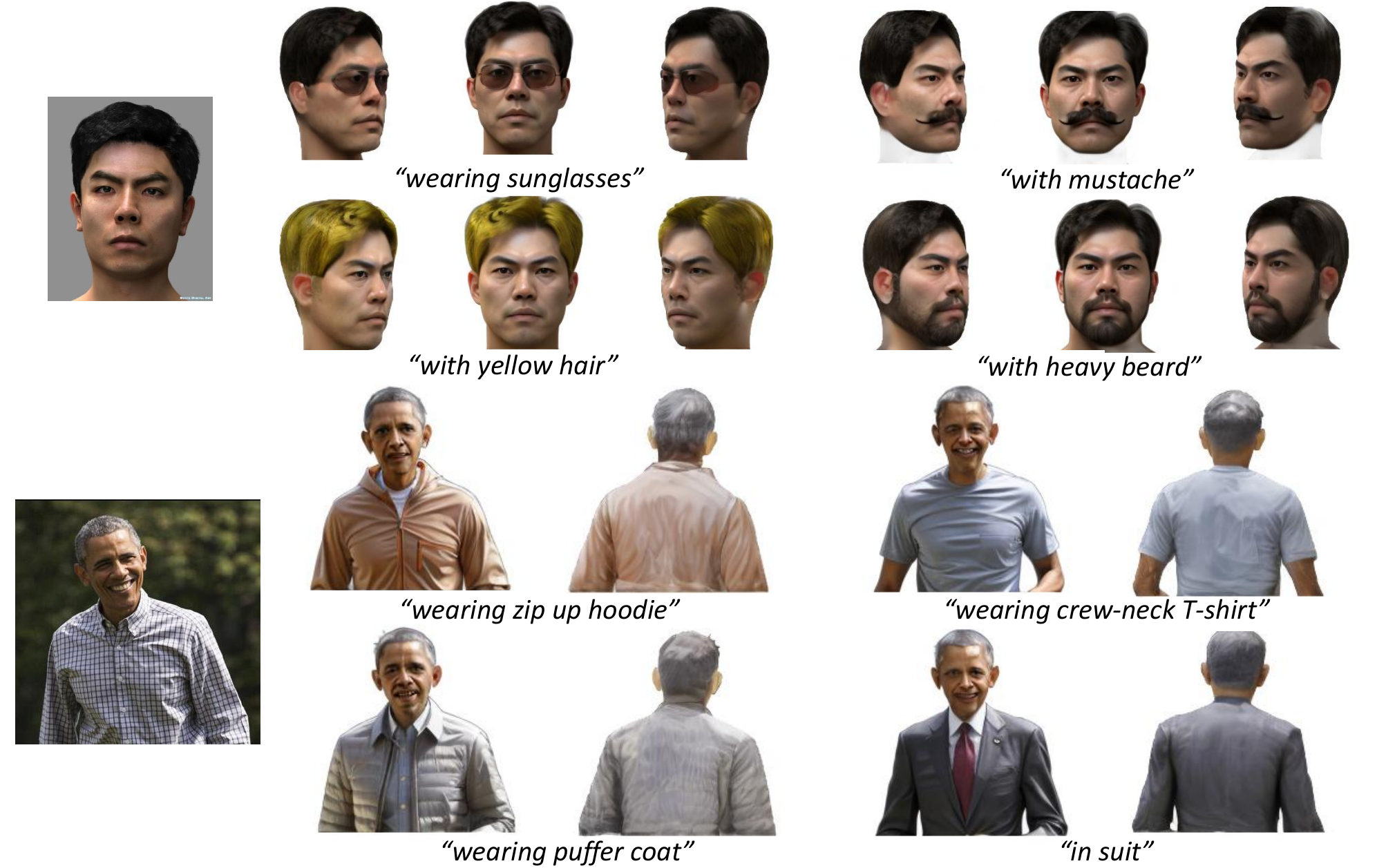}
    \caption{\textbf{More personalization results for humans.} Given a customized description and a face image, we can generate high-quality attributes (\eg, beard, clothes) for the 3D character according to various contexts.}
    \label{fig:human_customize_cases}
\end{figure*}


\begin{figure*}[h!]
    \centering
    \includegraphics[width=\linewidth]{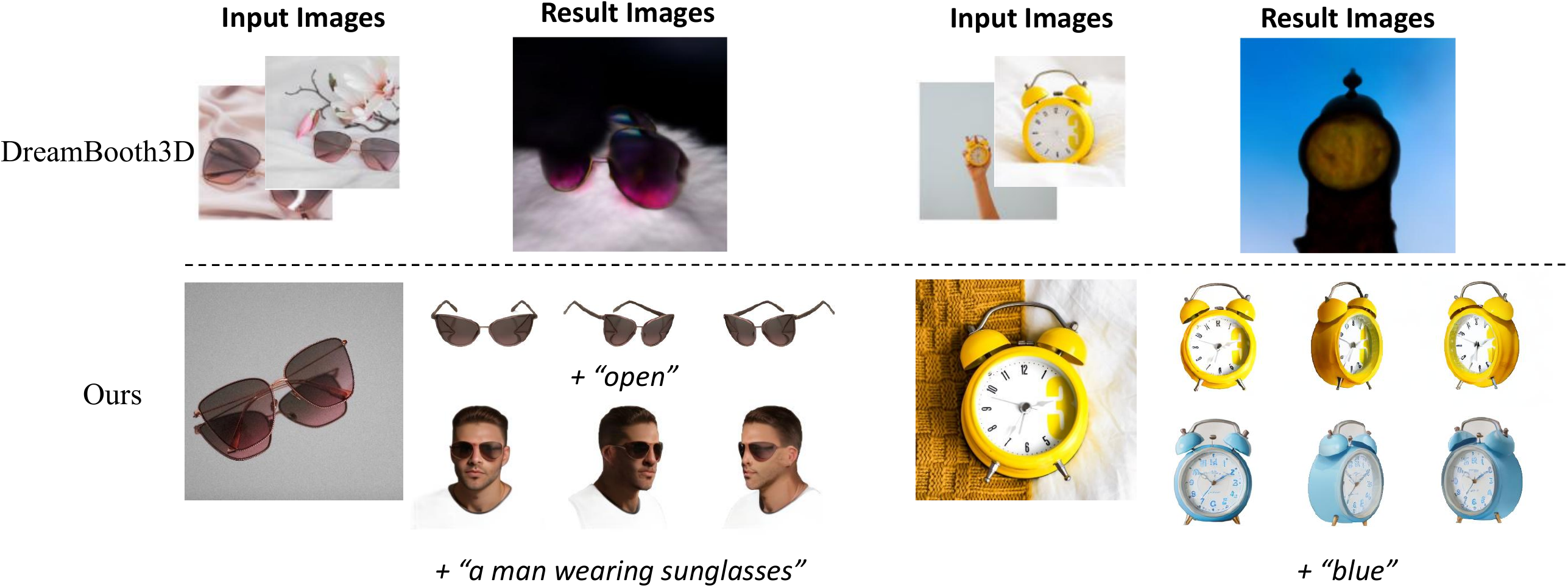}
    \caption{\textbf{Comparisons with the failure cases in DreamBooth3D~\cite{raj2023dreambooth3d}}. As Dreambooth3D fails to reconstruct thin object structures like sunglasses and suffers from limited view variation, Our method has made significant improvements in fine details of thin objects and fast 3D personalization from a single subject image.}
    \label{fig:dreambooth3d_failure_cases}
\end{figure*}
We carry out ablation studies on the design of Make-Your-3D framework in Fig.~\ref{fig:ablation_study} to verify the effectiveness of our co-evolution framework. Specifically, we perform ablation on identity-aware optimization and subject-prior optimization. The results reveal that the omission of any of two elements leads to a degradation in terms of subject-driven fidelity. Notably, the absence of identity-aware optimization leads to worse subject preservation and consistency. The lack of subject-prior optimization results in less plausible multi-view rendering, especially in cases where the back view lacks informative subject-prior guidance. This illustrates the 
effectiveness of our overall framework (Fig.~\ref{fig:pipeline}) that can approximate subject
\begin{wraptable}{r}{8.8cm}
    \centering
    \small
    \vspace{-3.5mm}
    \caption{\textbf{Quantitative comparison results} of DreamBooth3D~\cite{raj2023dreambooth3d}, multi-view DreamBooth~\cite{shi2023mvdream} and our Make-Your-3D on the multi-view consistency, subject fidelity, prompt fidelity, and overall quality score in a user study, rated on a range of 1-10, with higher scores indicating better performance.}
    \label{tab:user_study}
    \vspace{1.5mm}
    \begin{tabular}{lcccc}
        \toprule
        \multirow{2}*{Method}  & Multi-view & Subject & Prompt & Overall \\
                         & Consistency & Fidelity & Fidelity & Quality \\
        \midrule
        DreamBooth3D~\cite{raj2023dreambooth3d}         & 6.05   &  6.42  & 6.89   & 5.33    \\
        MV DreamBooth~\cite{shi2023mvdream}         &  8.76 &  7.55  &  6.73  &  7.59   \\
        Make-Your-3D (Ours)    & \textbf{9.01}  & \textbf{8.91} &  \textbf{8.70}  & \textbf{9.05}     \\     
        \bottomrule
    \end{tabular}
    \vspace{-7mm}
\end{wraptable}
distribution and have great identity-specific preservation. Moreover, our method is robust in various open-vocabulary settings from wild web images and achieves high-quality results in failure cases of DreamBooth3D~\cite{raj2023dreambooth3d} shown in Fig.~\ref{fig:dreambooth3d_failure_cases}. Driven by our co-evolution framework, we can serve more applications such as human personalization shown in Fig.~\ref{fig:human_customize_cases}, where we can change their attributes like hair, clothes, and more. These surprising results further support the 
effectiveness of the co-evolution framework in our Make-Your-3D and present the great potential for subject-driven customization. More impressive results on different applications can be found in our supplementary materials.

\section{Conclusion}

In this paper, we have proposed Make-Your-3D, a method for fast and consistent subject-driven 3D content generation. To approximate the distribution of the 3D subject, we introduce a novel co-evolution framework. This includes an identity-aware optimization for 2D personalized model and a subject-specific optimization for multi-view diffusion model, through which each model adapts and improves the other's capacity to capture the subject-driven identity. Therefore, our method bridges the distribution variance from the 3D subject, achieving high-fidelity, multi-view coherent, and subject-specific 3D assets that faithfully adhere to the contextualization in text guidance (\eg, playing guitar, boxing, etc.). Notably, we only need a single subject image as input and produce per 3D result within 5 minutes, $36 \times$ faster than DreamBooth3D~\cite{raj2023dreambooth3d}. Extensive qualitative and quantitative experiments verify the effectiveness and efficiency of our co-evolution framework on 3D content personalization and demonstrate the potential for a wide range of applications.
\newline

\noindent \textbf{Limitations and Future Work.} 
Although our Make-Your-3D allows for high-quality 3D personalization and demonstrates better performance than previous work, the quality still seems to be limited to the backbone itself based on Stable Diffusion v1.5. The larger diffusion model such as SDXL~\cite{SDXL} will further improve our performance. In future work, we are interested in exploring the 3D scene-level personalization which is a more challenging and complex task.   
We hope that our Make-Your-3D will pave the way for future advancements, as we believe this technology of subject-driven 3D generation may have a disruptive effect on various sectors, including advertising, entertainment, fashion, and more.

\bibliographystyle{splncs04}
\bibliography{egbib}

\begin{thebibliography}{10}
\providecommand{\url}[1]{\texttt{#1}}
\providecommand{\urlprefix}{URL }
\providecommand{\doi}[1]{https://doi.org/#1}

\bibitem{SDXL}
stable-diffusion-xl-base-1.0. \url{https://huggingface.co/stabilityai/stable-diffusion-xl-base-1.0}, accessed: 2023-08-29

\bibitem{achiam2023gpt-4v}
Achiam, J., Adler, S., Agarwal, S., Ahmad, L., Akkaya, I., Aleman, F.L., Almeida, D., Altenschmidt, J., Altman, S., Anadkat, S., et~al.: Gpt-4 technical report. arXiv preprint arXiv:2303.08774  (2023)

\bibitem{avrahami2023chosen-one}
Avrahami, O., Hertz, A., Vinker, Y., Arar, M., Fruchter, S., Fried, O., Cohen-Or, D., Lischinski, D.: The chosen one: Consistent characters in text-to-image diffusion models. arXiv preprint arXiv:2311.10093  (2023)

\bibitem{10419041}
Cao, H., Tan, C., Gao, Z., Xu, Y., Chen, G., Heng, P.A., Li, S.Z.: A survey on generative diffusion models. IEEE Transactions on Knowledge and Data Engineering pp. 1--20 (2024). \doi{10.1109/TKDE.2024.3361474}

\bibitem{chang2015shapenet}
Chang, A.X., Funkhouser, T., Guibas, L., Hanrahan, P., Huang, Q., Li, Z., Savarese, S., Savva, M., Song, S., Su, H., et~al.: Shapenet: An information-rich 3d model repository. arXiv preprint arXiv:1512.03012  (2015)

\bibitem{chen2023singlestage}
Chen, H., Gu, J., Chen, A., Tian, W., Tu, Z., Liu, L., Su, H.: Single-stage diffusion nerf: A unified approach to 3d generation and reconstruction (2023)

\bibitem{chen2023fantasia3d}
Chen, R., Chen, Y., Jiao, N., Jia, K.: Fantasia3d: Disentangling geometry and appearance for high-quality text-to-3d content creation (2023)

\bibitem{chen2024subject-apprenticeship}
Chen, W., Hu, H., Li, Y., Ruiz, N., Jia, X., Chang, M.W., Cohen, W.W.: Subject-driven text-to-image generation via apprenticeship learning. Advances in Neural Information Processing Systems  \textbf{36} (2024)

\bibitem{deitke2024objaverse-xl}
Deitke, M., Liu, R., Wallingford, M., Ngo, H., Michel, O., Kusupati, A., Fan, A., Laforte, C., Voleti, V., Gadre, S.Y., et~al.: Objaverse-xl: A universe of 10m+ 3d objects. Advances in Neural Information Processing Systems  \textbf{36} (2024)

\bibitem{deitke2023objaverse}
Deitke, M., Schwenk, D., Salvador, J., Weihs, L., Michel, O., VanderBilt, E., Schmidt, L., Ehsani, K., Kembhavi, A., Farhadi, A.: Objaverse: A universe of annotated 3d objects. In: Proceedings of the IEEE/CVF Conference on Computer Vision and Pattern Recognition. pp. 13142--13153 (2023)

\bibitem{deng2022nerdi}
Deng, C., Jiang, C.M., Qi, C.R., Yan, X., Zhou, Y., Guibas, L., Anguelov, D.: Nerdi: Single-view nerf synthesis with language-guided diffusion as general image priors (2022)

\bibitem{du2024stable-is-unstable}
Du, C., Li, Y., Qiu, Z., Xu, C.: Stable diffusion is unstable. Advances in Neural Information Processing Systems  \textbf{36} (2024)

\bibitem{eftekhar2021omnidata-normal-estimator}
Eftekhar, A., Sax, A., Malik, J., Zamir, A.: Omnidata: A scalable pipeline for making multi-task mid-level vision datasets from 3d scans. In: Proceedings of the IEEE/CVF International Conference on Computer Vision. pp. 10786--10796 (2021)

\bibitem{gal2022image}
Gal, R., Alaluf, Y., Atzmon, Y., Patashnik, O., Bermano, A.H., Chechik, G., Cohen-Or, D.: An image is worth one word: Personalizing text-to-image generation using textual inversion (2022)

\bibitem{gupta20233dgen}
Gupta, A., Xiong, W., Nie, Y., Jones, I., Oğuz, B.: 3dgen: Triplane latent diffusion for textured mesh generation (2023)

\bibitem{ho2020ddpm}
Ho, J., Jain, A., Abbeel, P.: Denoising diffusion probabilistic models. Advances in neural information processing systems  \textbf{33},  6840--6851 (2020)

\bibitem{ho2020denoising}
Ho, J., Jain, A., Abbeel, P.: Denoising diffusion probabilistic models (2020)

\bibitem{ho2022classifier-free-guidance}
Ho, J., Salimans, T.: Classifier-free diffusion guidance. arXiv preprint arXiv:2207.12598  (2022)

\bibitem{huang2024customizeit3d}
Huang, N., Zhang, T., Yuan, Y., Chen, D., Zhang, S.: Customize-it-3d: High-quality 3d creation from a single image using subject-specific knowledge prior (2024)

\bibitem{jiang2023videobooth}
Jiang, Y., Wu, T., Yang, S., Si, C., Lin, D., Qiao, Y., Loy, C.C., Liu, Z.: Videobooth: Diffusion-based video generation with image prompts (2023)

\bibitem{kerbl20233d-gaussian}
Kerbl, B., Kopanas, G., Leimk{\"u}hler, T., Drettakis, G.: 3d gaussian splatting for real-time radiance field rendering. ACM Transactions on Graphics  \textbf{42}(4) (2023)

\bibitem{kim2023neuralfieldldm}
Kim, S.W., Brown, B., Yin, K., Kreis, K., Schwarz, K., Li, D., Rombach, R., Torralba, A., Fidler, S.: Neuralfield-ldm: Scene generation with hierarchical latent diffusion models (2023)

\bibitem{kumari2023multiconcept}
Kumari, N., Zhang, B., Zhang, R., Shechtman, E., Zhu, J.Y.: Multi-concept customization of text-to-image diffusion (2023)

\bibitem{li2023generative}
Li, C., Zhang, C., Waghwase, A., Lee, L.H., Rameau, F., Yang, Y., Bae, S.H., Hong, C.S.: Generative ai meets 3d: A survey on text-to-3d in aigc era. arXiv preprint arXiv:2305.06131  (2023)

\bibitem{li2024advances}
Li, X., Zhang, Q., Kang, D., Cheng, W., Gao, Y., Zhang, J., Liang, Z., Liao, J., Cao, Y.P., Shan, Y.: Advances in 3d generation: A survey (2024)

\bibitem{lin2023magic3d}
Lin, C.H., Gao, J., Tang, L., Takikawa, T., Zeng, X., Huang, X., Kreis, K., Fidler, S., Liu, M.Y., Lin, T.Y.: Magic3d: High-resolution text-to-3d content creation. In: Proceedings of the IEEE/CVF Conference on Computer Vision and Pattern Recognition. pp. 300--309 (2023)

\bibitem{liu2023sherpa3d}
Liu, F., Wu, D., Wei, Y., Rao, Y., Duan, Y.: Sherpa3d: Boosting high-fidelity text-to-3d generation via coarse 3d prior (2023)

\bibitem{liu2024comprehensive}
Liu, J., Huang, X., Huang, T., Chen, L., Hou, Y., Tang, S., Liu, Z., Ouyang, W., Zuo, W., Jiang, J., et~al.: A comprehensive survey on 3d content generation. arXiv preprint arXiv:2402.01166  (2024)

\bibitem{long2023wonder3d}
Long, X., Guo, Y.C., Lin, C., Liu, Y., Dou, Z., Liu, L., Ma, Y., Zhang, S.H., Habermann, M., Theobalt, C., et~al.: Wonder3d: Single image to 3d using cross-domain diffusion. arXiv preprint arXiv:2310.15008  (2023)

\bibitem{lorraine2023att3d}
Lorraine, J., Xie, K., Zeng, X., Lin, C.H., Takikawa, T., Sharp, N., Lin, T.Y., Liu, M.Y., Fidler, S., Lucas, J.: Att3d: Amortized text-to-3d object synthesis (2023)

\bibitem{luo2021diffusion}
Luo, S., Hu, W.: Diffusion probabilistic models for 3d point cloud generation (2021)

\bibitem{ma2024magic-me}
Ma, Z., Zhou, D., Yeh, C.H., Wang, X.S., Li, X., Yang, H., Dong, Z., Keutzer, K., Feng, J.: Magic-me: Identity-specific video customized diffusion. arXiv preprint arXiv:2402.09368  (2024)

\bibitem{melaskyriazi2023realfusion}
Melas-Kyriazi, L., Rupprecht, C., Laina, I., Vedaldi, A.: Realfusion: 360${\deg}$ reconstruction of any object from a single image (2023)

\bibitem{mildenhall2021nerf-eccv}
Mildenhall, B., Srinivasan, P.P., Tancik, M., Barron, J.T., Ramamoorthi, R., Ng, R.: Nerf: Representing scenes as neural radiance fields for view synthesis. Communications of the ACM  \textbf{65}(1),  99--106 (2021)

\bibitem{muller2022instant-NGP}
M{\"u}ller, T., Evans, A., Schied, C., Keller, A.: Instant neural graphics primitives with a multiresolution hash encoding. ACM Transactions on Graphics (ToG)  \textbf{41}(4),  1--15 (2022)

\bibitem{nichol2022pointe}
Nichol, A., Jun, H., Dhariwal, P., Mishkin, P., Chen, M.: Point-e: A system for generating 3d point clouds from complex prompts (2022)

\bibitem{poole2022dreamfusion}
Poole, B., Jain, A., Barron, J.T., Mildenhall, B.: Dreamfusion: Text-to-3d using 2d diffusion (2022)

\bibitem{qian2023magic123}
Qian, G., Mai, J., Hamdi, A., Ren, J., Siarohin, A., Li, B., Lee, H.Y., Skorokhodov, I., Wonka, P., Tulyakov, S., Ghanem, B.: Magic123: One image to high-quality 3d object generation using both 2d and 3d diffusion priors (2023)

\bibitem{radford2021-CLIP}
Radford, A., Kim, J.W., Hallacy, C., Ramesh, A., Goh, G., Agarwal, S., Sastry, G., Askell, A., Mishkin, P., Clark, J., et~al.: Learning transferable visual models from natural language supervision. In: International conference on machine learning. pp. 8748--8763. PMLR (2021)

\bibitem{raj2023dreambooth3d}
Raj, A., Kaza, S., Poole, B., Niemeyer, M., Ruiz, N., Mildenhall, B., Zada, S., Aberman, K., Rubinstein, M., Barron, J., et~al.: Dreambooth3d: Subject-driven text-to-3d generation. arXiv preprint arXiv:2303.13508  (2023)

\bibitem{reed2016generative}
Reed, S., Akata, Z., Yan, X., Logeswaran, L., Schiele, B., Lee, H.: Generative adversarial text to image synthesis. In: International conference on machine learning. pp. 1060--1069. PMLR (2016)

\bibitem{ren2024customizeavideo}
Ren, Y., Zhou, Y., Yang, J., Shi, J., Liu, D., Liu, F., Kwon, M., Shrivastava, A.: Customize-a-video: One-shot motion customization of text-to-video diffusion models (2024)

\bibitem{rombach2022stable-diffusion}
Rombach, R., Blattmann, A., Lorenz, D., Esser, P., Ommer, B.: High-resolution image synthesis with latent diffusion models. In: Proceedings of the IEEE/CVF conference on computer vision and pattern recognition. pp. 10684--10695 (2022)

\bibitem{ruiz2023dreambooth}
Ruiz, N., Li, Y., Jampani, V., Pritch, Y., Rubinstein, M., Aberman, K.: Dreambooth: Fine tuning text-to-image diffusion models for subject-driven generation. In: Proceedings of the IEEE/CVF Conference on Computer Vision and Pattern Recognition. pp. 22500--22510 (2023)

\bibitem{ruiz2023hyperdreambooth}
Ruiz, N., Li, Y., Jampani, V., Wei, W., Hou, T., Pritch, Y., Wadhwa, N., Rubinstein, M., Aberman, K.: Hyperdreambooth: Hypernetworks for fast personalization of text-to-image models. arXiv preprint arXiv:2307.06949  (2023)

\bibitem{saharia2022photorealistic-imagen}
Saharia, C., Chan, W., Saxena, S., Li, L., Whang, J., Denton, E., Ghasemipour, S.K.S., Ayan, B.K., Mahdavi, S.S., Lopes, R.G., Salimans, T., Ho, J., Fleet, D.J., Norouzi, M.: Photorealistic text-to-image diffusion models with deep language understanding (2022)

\bibitem{shi2023mvdream}
Shi, Y., Wang, P., Ye, J., Long, M., Li, K., Yang, X.: Mvdream: Multi-view diffusion for 3d generation (2023)

\bibitem{shue20223d}
Shue, J.R., Chan, E.R., Po, R., Ankner, Z., Wu, J., Wetzstein, G.: 3d neural field generation using triplane diffusion (2022)

\bibitem{sohl2015deep-diffusion}
Sohl-Dickstein, J., Weiss, E., Maheswaranathan, N., Ganguli, S.: Deep unsupervised learning using nonequilibrium thermodynamics. In: International conference on machine learning. pp. 2256--2265. PMLR (2015)

\bibitem{sun2023dreamcraft3d}
Sun, J., Zhang, B., Shao, R., Wang, L., Liu, W., Xie, Z., Liu, Y.: Dreamcraft3d: Hierarchical 3d generation with bootstrapped diffusion prior (2023)

\bibitem{tang2024lgm}
Tang, J., Chen, Z., Chen, X., Wang, T., Zeng, G., Liu, Z.: Lgm: Large multi-view gaussian model for high-resolution 3d content creation (2024)

\bibitem{tang2023dreamgaussian}
Tang, J., Ren, J., Zhou, H., Liu, Z., Zeng, G.: Dreamgaussian: Generative gaussian splatting for efficient 3d content creation. arXiv preprint arXiv:2309.16653  (2023)

\bibitem{tang2023delicate-nerf-to-mesh}
Tang, J., Zhou, H., Chen, X., Hu, T., Ding, E., Wang, J., Zeng, G.: Delicate textured mesh recovery from nerf via adaptive surface refinement. arXiv preprint arXiv:2303.02091  (2023)

\bibitem{tang2023makeit3d}
Tang, J., Wang, T., Zhang, B., Zhang, T., Yi, R., Ma, L., Chen, D.: Make-it-3d: High-fidelity 3d creation from a single image with diffusion prior (2023)

\bibitem{tang2023mvdiffusion}
Tang, S., Zhang, F., Chen, J., Wang, P., Furukawa, Y.: Mvdiffusion: Enabling holistic multi-view image generation with correspondence-aware diffusion (2023)

\bibitem{wang2023imagedream}
Wang, P., Shi, Y.: Imagedream: Image-prompt multi-view diffusion for 3d generation. arXiv preprint arXiv:2312.02201  (2023)

\bibitem{wang2023prolificdreamer}
Wang, Z., Lu, C., Wang, Y., Bao, F., Li, C., Su, H., Zhu, J.: Prolificdreamer: High-fidelity and diverse text-to-3d generation with variational score distillation (2023)

\bibitem{wei2023dreamvideo}
Wei, Y., Zhang, S., Qing, Z., Yuan, H., Liu, Z., Liu, Y., Zhang, Y., Zhou, J., Shan, H.: Dreamvideo: Composing your dream videos with customized subject and motion. arXiv preprint arXiv:2312.04433  (2023)

\bibitem{woo2023harmonyview}
Woo, S., Park, B., Go, H., Kim, J.Y., Kim, C.: Harmonyview: Harmonizing consistency and diversity in one-image-to-3d (2023)

\bibitem{wu2023tune-a-video}
Wu, J.Z., Ge, Y., Wang, X., Lei, S.W., Gu, Y., Shi, Y., Hsu, W., Shan, Y., Qie, X., Shou, M.Z.: Tune-a-video: One-shot tuning of image diffusion models for text-to-video generation. In: Proceedings of the IEEE/CVF International Conference on Computer Vision. pp. 7623--7633 (2023)

\bibitem{wu2023aigc-survey}
Wu, J., Gan, W., Chen, Z., Wan, S., Lin, H.: Ai-generated content (aigc): A survey. arXiv preprint arXiv:2304.06632  (2023)

\bibitem{wu2024gpt4v-metric}
Wu, T., Yang, G., Li, Z., Zhang, K., Liu, Z., Guibas, L., Lin, D., Wetzstein, G.: Gpt-4v (ision) is a human-aligned evaluator for text-to-3d generation. arXiv preprint arXiv:2401.04092  (2024)

\bibitem{xing2023make-your-video}
Xing, J., Xia, M., Liu, Y., Zhang, Y., Zhang, Y., He, Y., Liu, H., Chen, H., Cun, X., Wang, X., et~al.: Make-your-video: Customized video generation using textual and structural guidance. arXiv preprint arXiv:2306.00943  (2023)

\bibitem{xing2023survey}
Xing, Z., Feng, Q., Chen, H., Dai, Q., Hu, H., Xu, H., Wu, Z., Jiang, Y.G.: A survey on video diffusion models (2023)

\bibitem{xu2023neurallift360}
Xu, D., Jiang, Y., Wang, P., Fan, Z., Wang, Y., Wang, Z.: Neurallift-360: Lifting an in-the-wild 2d photo to a 3d object with 360${\deg}$ views (2023)

\bibitem{yang2023diffusion-survey}
Yang, L., Zhang, Z., Song, Y., Hong, S., Xu, R., Zhao, Y., Zhang, W., Cui, B., Yang, M.H.: Diffusion models: A comprehensive survey of methods and applications. ACM Computing Surveys  \textbf{56}(4),  1--39 (2023)

\bibitem{ye2023ip-adapter}
Ye, H., Zhang, J., Liu, S., Han, X., Yang, W.: Ip-adapter: Text compatible image prompt adapter for text-to-image diffusion models. arXiv preprint arXiv:2308.06721  (2023)

\bibitem{zeng2023avatarbooth}
Zeng, Y., Lu, Y., Ji, X., Yao, Y., Zhu, H., Cao, X.: Avatarbooth: High-quality and customizable 3d human avatar generation. arXiv preprint arXiv:2306.09864  (2023)

\bibitem{zhang20233dshape2vecset}
Zhang, B., Tang, J., Niessner, M., Wonka, P.: 3dshape2vecset: A 3d shape representation for neural fields and generative diffusion models (2023)

\bibitem{zhang2023text-diffusion-survey}
Zhang, C., Zhang, C., Zhang, M., Kweon, I.S.: Text-to-image diffusion model in generative ai: A survey. arXiv preprint arXiv:2303.07909  (2023)

\bibitem{zhang2023texttoimage}
Zhang, C., Zhang, C., Zhang, M., Kweon, I.S.: Text-to-image diffusion models in generative ai: A survey (2023)

\bibitem{zhang2023repaint123}
Zhang, J., Tang, Z., Pang, Y., Cheng, X., Jin, P., Wei, Y., Ning, M., Yuan, L.: Repaint123: Fast and high-quality one image to 3d generation with progressive controllable 2d repainting (2023)

\bibitem{zhao2023michelangelo}
Zhao, Z., Liu, W., Chen, X., Zeng, X., Wang, R., Cheng, P., Fu, B., Chen, T., Yu, G., Gao, S.: Michelangelo: Conditional 3d shape generation based on shape-image-text aligned latent representation (2023)

\end{thebibliography}

\appendix

\thispagestyle{empty}

\section{Additional Implementation Details}
In this section, we provide additional details about the implementation of our subject-driven 3D generation framework \textbf{Make-Your-3D}. 

\subsection{Hyperparameter Settings}
In our subject-driven optimization, We retain the optimizer settings and $\epsilon$-prediction strategy from the pretrained process, with a $0.9$ adam $\beta_1$, a $0.999$ adam $\beta_2$, a $1e-2$ adam weight decay, and a $1e-8$ adam $\epsilon$. During the optimization, we use a reduced image size of 256$\times$256.

\subsection{Training Details}
The time consumption for the three stages in our framework is as follows: identity-aware optimization takes $\sim$ 1 minute, subject-driven optimization takes $\sim$ 3 minutes, and mesh conversion takes approximately $\sim$ 1 minute. The mesh conversion module is designed to be adaptable to a variety of mesh extraction models~\cite{tang2023delicate-nerf-to-mesh, long2023wonder3d, tang2024lgm}. For the sake of improved efficiency, we have chosen to utilize the LGM~\cite{tang2024lgm}, which is capable of generating 3D Gaussians of objects within a mere 7 seconds. After that, we train an efficient NeRF (\ie, Instant-NGP~\cite{muller2022instant-NGP}) by using the rendered images from 3D Gaussians, and then convert the NeRF to polygonal meshes~\cite{tang2023delicate-nerf-to-mesh}. Specifically, we train two hash grids to reconstruct the geometry and appearance from Gaussian renderings. Please refer to LGM~\cite{tang2024lgm} for more details of generated Gaussians. With adequately optimized implementation, it takes extra 1 minute to perform this Gaussians to NeRF to mesh conversion. Our codes for
implementation will be available upon acceptance.

\subsection{Designing Prompts for One-Shot Personalization}
Our goal is to let the diffusion model’s be deeply aware of a new subject's identity. As mentioned in DreamBooth~\cite{ruiz2023dreambooth}, a common way to personalize a diffusion model is to “implant” a new (\textit{unique identifier, subject}) pair into the model’s “dictionary” and  label input multi-view images of the subject "a [identifier] [class noun]", where [identifier] is a unique identifier (\eg, “xxy5syt00”) linked to the subject and [class noun] is a coarse class descriptor of the subject (\eg, cat, dog, watch, etc.). However, this simple method loses all the position information in multi-view images, which is strongly important for identity awareness. To fully leverage the concealed information, we propose a novel \textit{(unique identifier, subject, direction)} pair inspired by view-dependent prompting in DreamFusion~\cite{poole2022dreamfusion} and automatically label each image "a [identifier] [class noun], [direction]", where "direction" is one of the six directions (\eg, front, back, left, etc.) for the corresponding multi-view image. By employing this design, we ensure that positional cues are incorporated into our identity-aware optimization process. We utilize the resulting six matched pairs for the optimization, further enhancing the model's ability to capture and comprehend the subject's identity.

\section{Additional Results}
To further demonstrate the effectiveness and impressive visualization results of our Make-Your-3D, we conducted more experiments including additional comparison results against DreamBooth3D~\cite{raj2023dreambooth3d} and visual results (\eg, multi-view images, textured meshes, normals, etc.).

\subsection{More Qualitative Comparisons}

\begin{figure*}[!h]
    \vspace{-0.3cm}
    \centering
    \includegraphics[width=\linewidth]{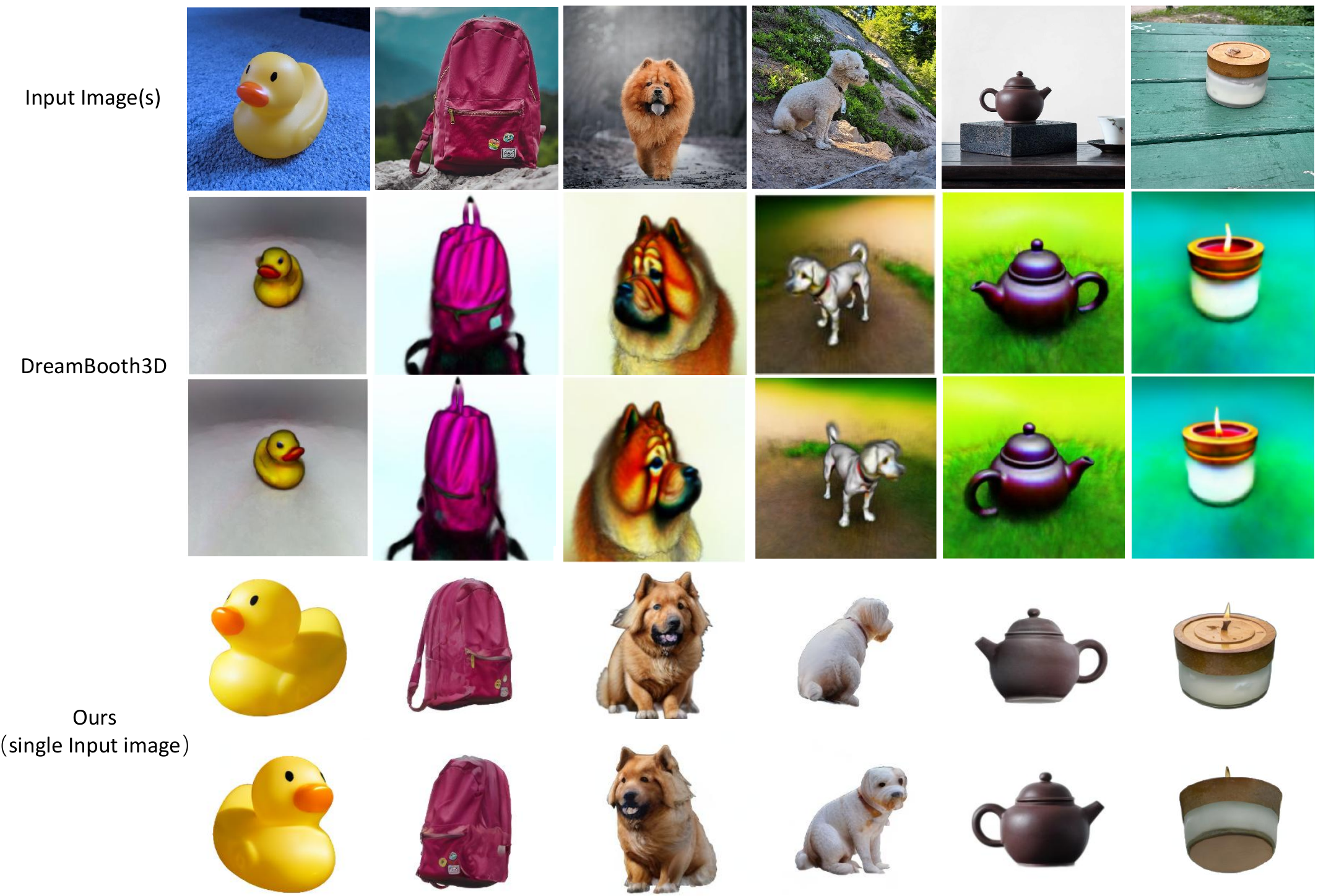}
    \caption{\textbf{More qualitative comparisons with DreamBooth3D} on subjects of DreamBooth dataset. Notice ours perform better on the object's identity consistency with less input images.}
    \label{fig:supplm_compare_with_dreambooth3d}
\end{figure*}

Fig.~\ref{fig:supplm_compare_with_dreambooth3d} demonstrates additional qualitative comparisons with DreamBooth3D~\cite{raj2023dreambooth3d}. We observed that DreamBooth3D tends to generate coarse results for limited subjects, which lack sufficient identity consistency and also suffer from overfitting issues. For example, the red backpack in Fig.~\ref{fig:supplm_compare_with_dreambooth3d} exhibits three small signs at the right bottom, which are not preserved by DreamBooth during its generation process. On the other hand, our proposed method successfully maintains this distinct feature while generating high-resolution outputs.

\subsection{More Visual Results}

\begin{figure*}[!t]
    \centering
    \includegraphics[width=\linewidth]{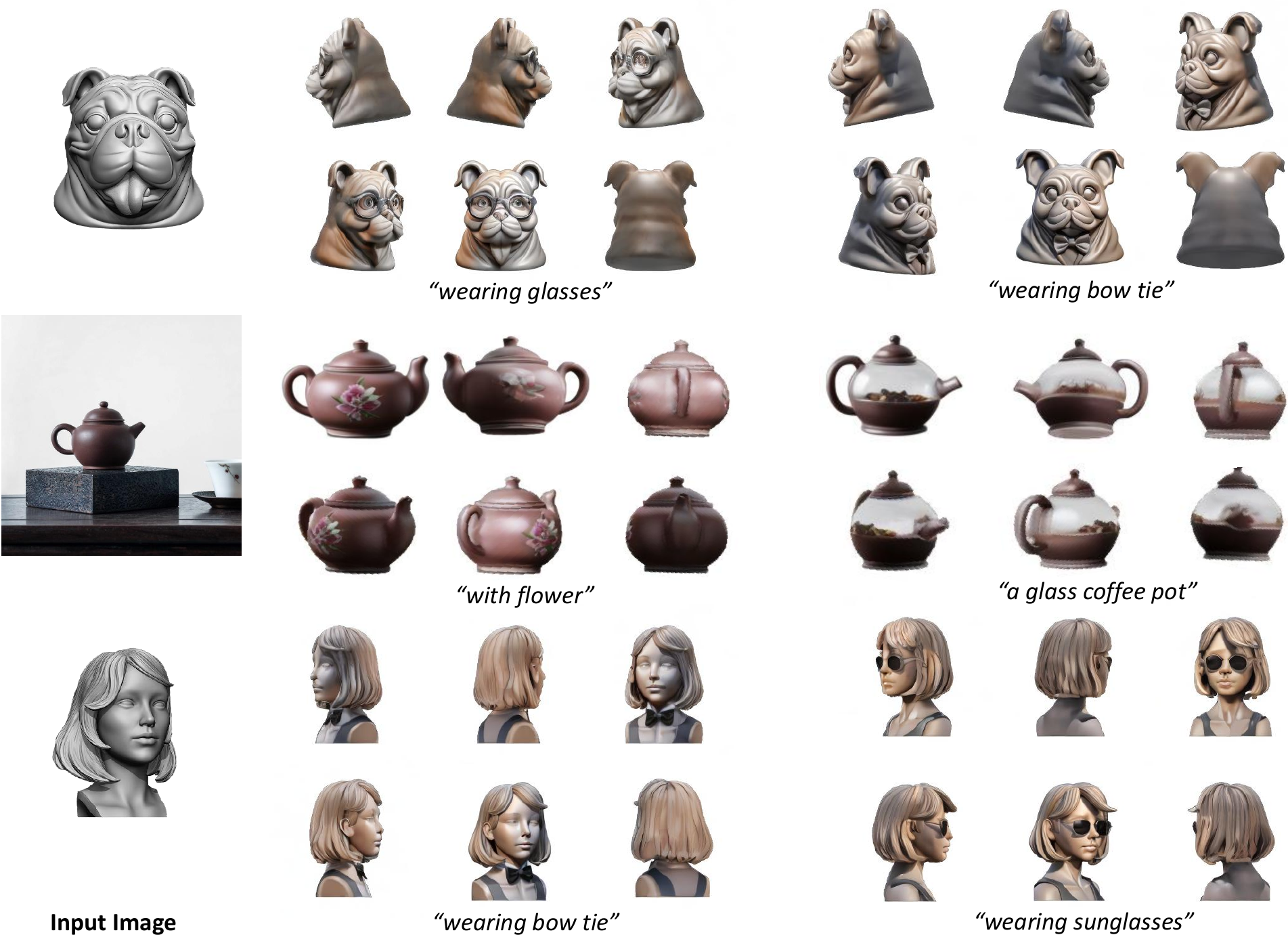}
    \caption{\textbf{More personalization results} for a subject with different text inputs.}
    \label{fig:supplm_prompt_diversity}
    \vspace{-0.3cm}
\end{figure*}

\begin{figure*}[!h]
    \centering
    \includegraphics[width=0.93\linewidth]{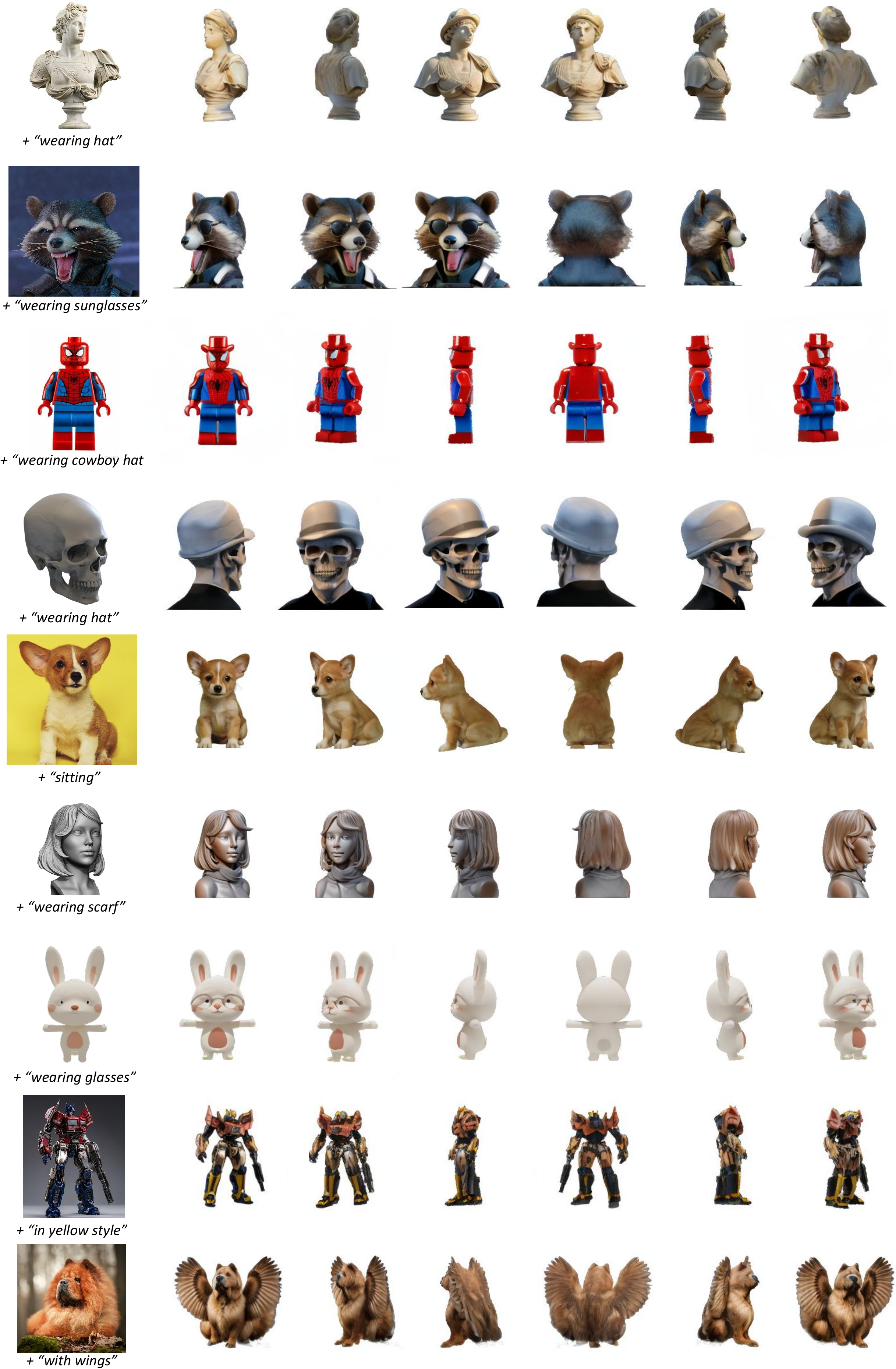}
    \caption{\textbf{More visual results of Make-Your-3D} on different subjects with customized text inputs.}
    \label{fig:supplm_multi_view_result}
\end{figure*}

\begin{figure*}[!t]
    \centering
    \includegraphics[width=\linewidth]{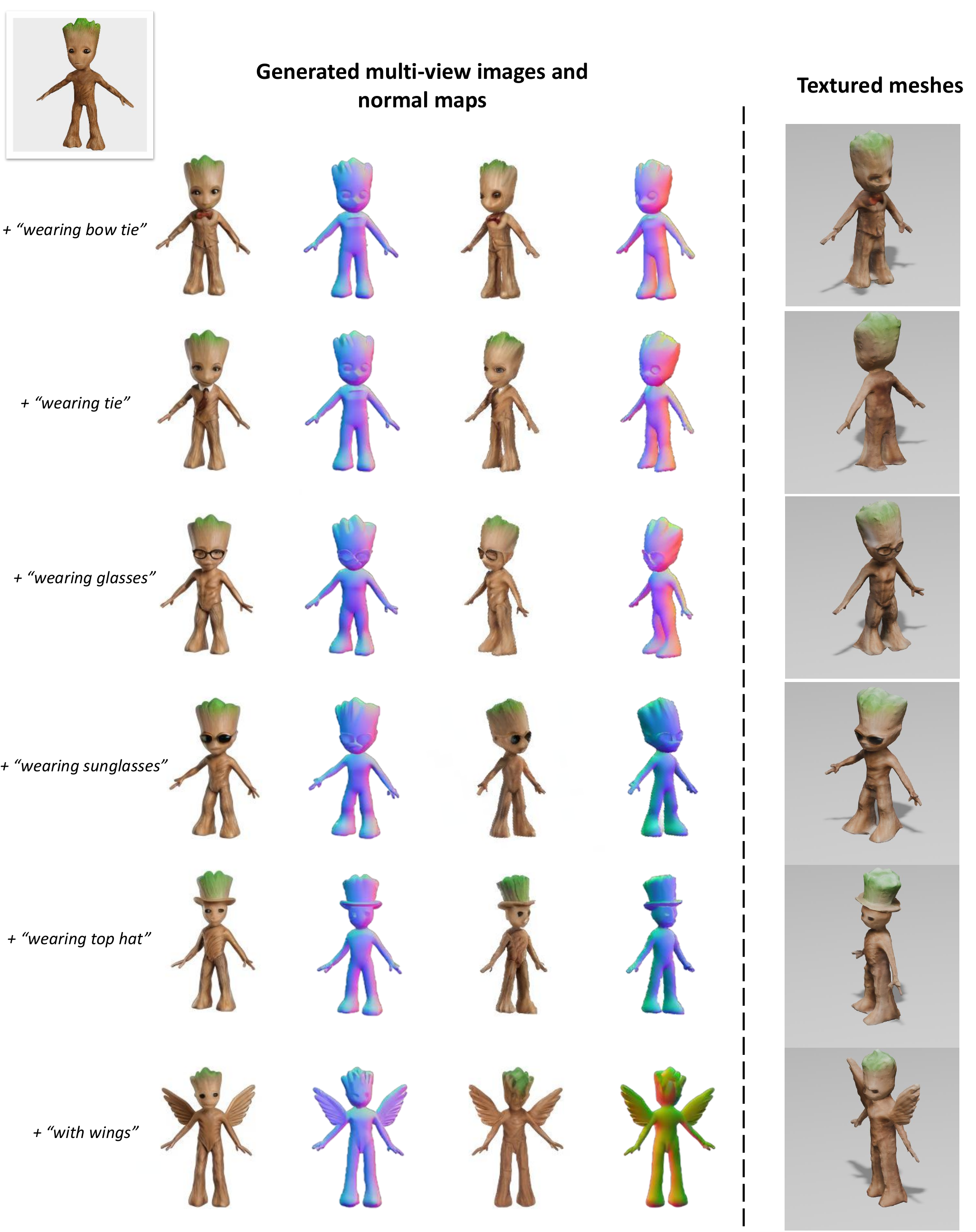}
    \caption{\textbf{Textured meshes and normal maps} on a subject "Gelute" with various customized text inputs.}
    \label{fig:supplm_textured_mesh}
\end{figure*}

Fig.~\ref{fig:supplm_prompt_diversity}, \ref{fig:supplm_multi_view_result} shows multiple views of the assets rendered for the subjects with different text prompts. Fig.~\ref{fig:supplm_textured_mesh} provides additional results with associated normals and textured meshes to demonstrate the 3D consistency of our results on a variety of customized subjects.

\subsection{More Human Personalization Results}
\begin{figure*}[!h]
    \vspace{0.2cm}
    \centering
    \includegraphics[width=\linewidth]{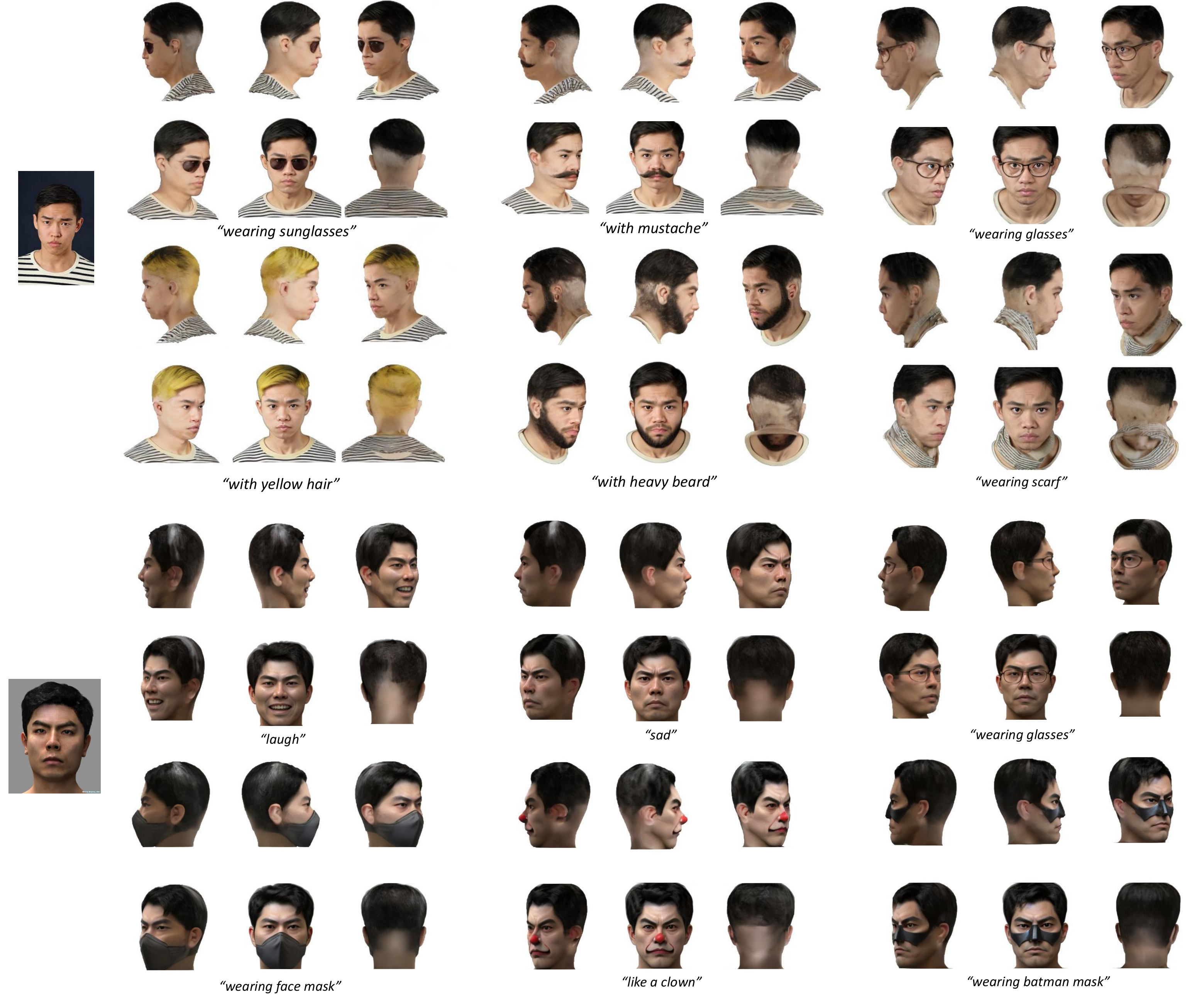}
    \caption{\textbf{More personalization results for human faces}.}
    \label{fig:supplm_human_face}
    \vspace{-0.2cm}
\end{figure*}

\begin{figure*}[!t]
    \centering
    \includegraphics[width=\linewidth]{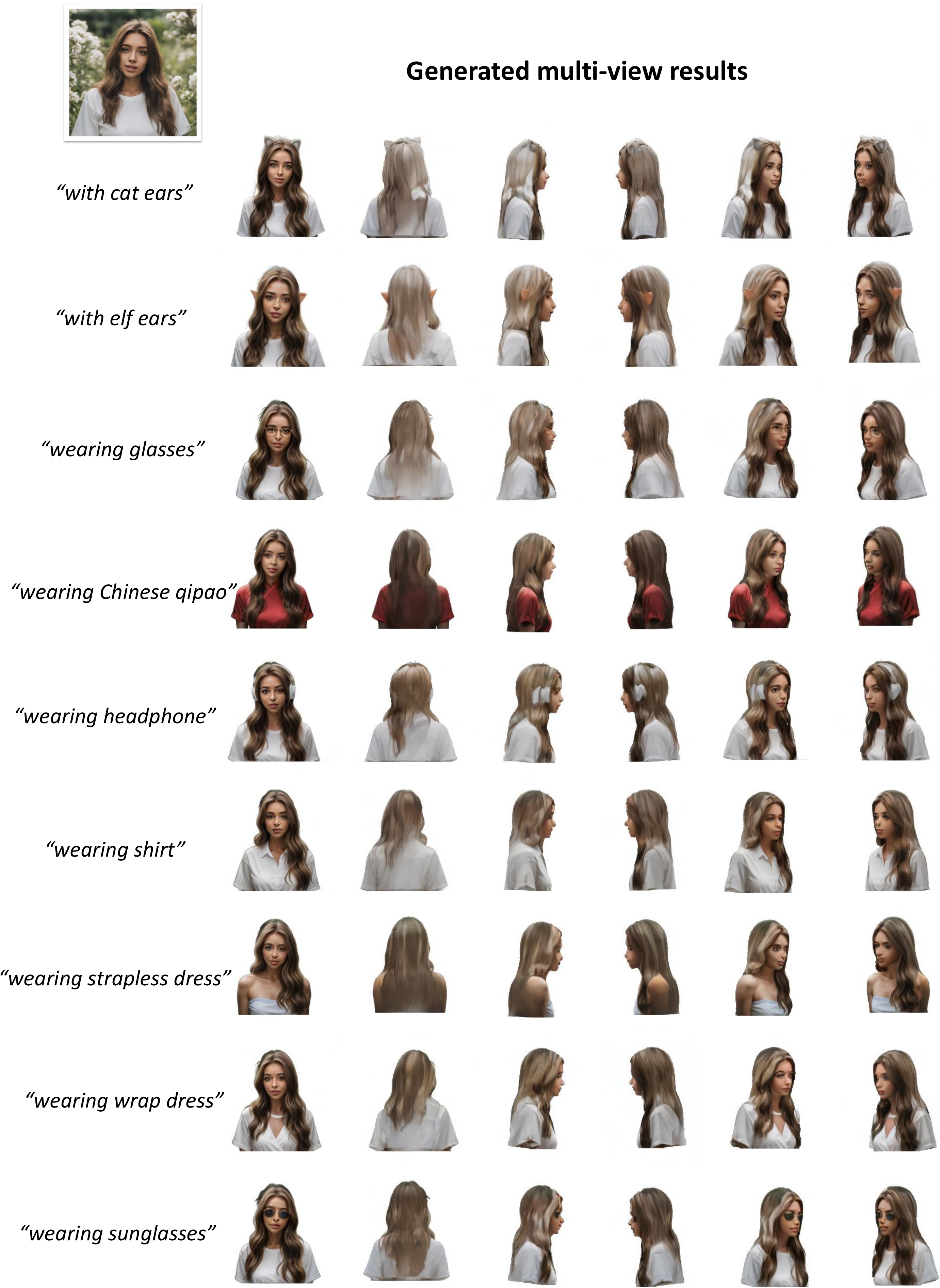}
    \caption{\textbf{More personalization results} for one person with multiple customized text inputs.}
    \label{fig:supplm_human_cloth}
\end{figure*}

Fig.~\ref{fig:supplm_human_face}, \ref{fig:supplm_human_cloth} presents additional personalized examples of human faces generated by our co-evolution framework. Our approach enables modification of attributes such as hairstyle, clothing, and more, as well as the capability to alter expressions, makeup, and styling. These remarkable results provide further evidence of the effectiveness of the co-evolution framework in our Make-Your-3D application and demonstrate its broad potential for various areas, such as customizing 3D characters, virtual reality, online clothing try-on, and beyond.



\section{GPT4-V for 3D Evaluation}
\begin{figure*}[!h]
    \vspace{0.4cm}
    \centering
    \includegraphics[width=\linewidth]{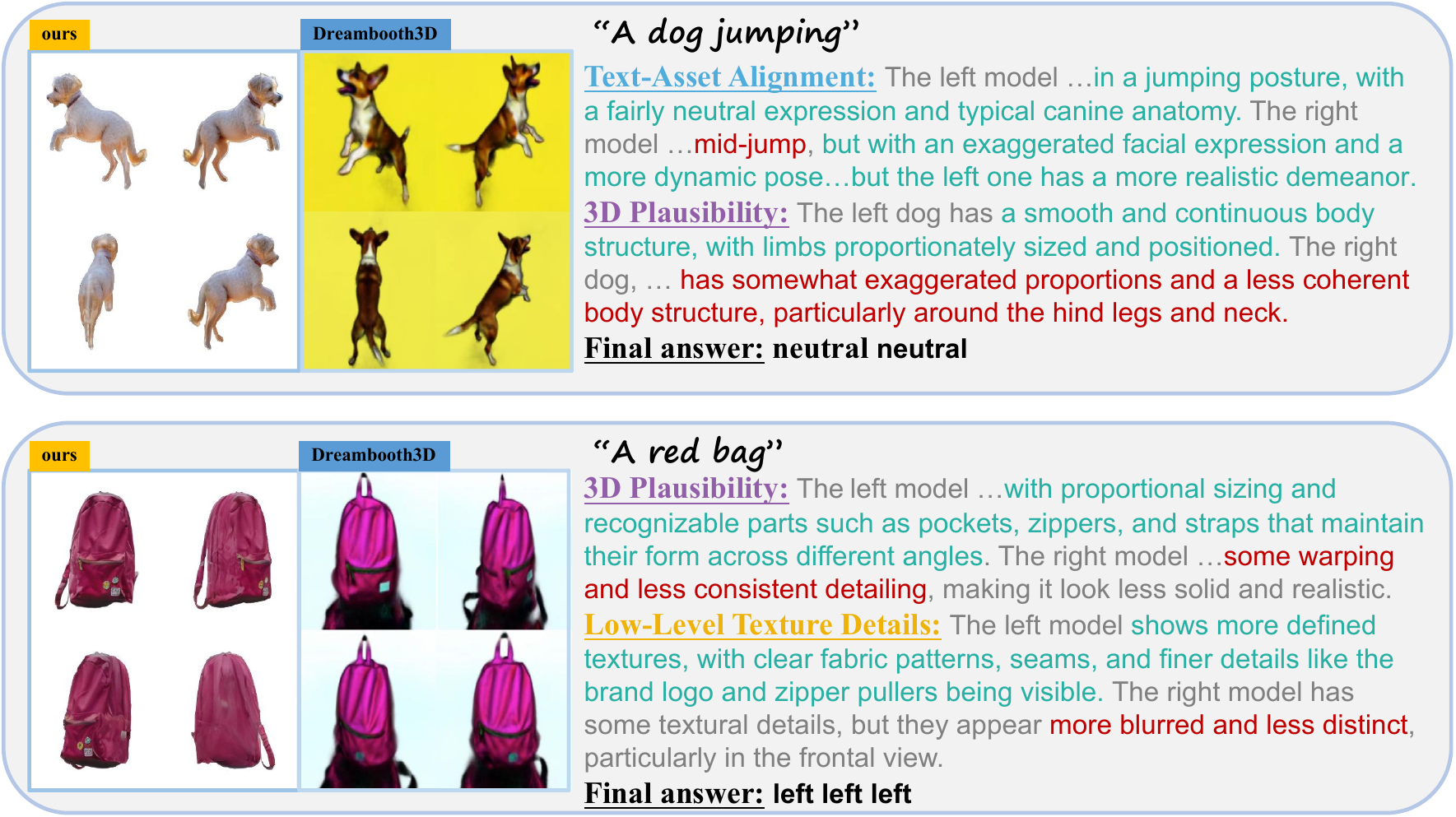}

    \caption{\textbf{Examples of the analysis by GPT-4V~\cite{achiam2023gpt-4v}.} Given two 3D assets, we ask GPT-4V to compare them on various aspects and provide an explanation. We find that GPT-4V’s preference closely aligns with that of humans in our user study.}
    \label{fig:supplm_gpt4}
\end{figure*}

We choose a recent automatic and versatile evaluation metric GPTEval3D~\cite{wu2024gpt4v-metric} based on GPT-4Vision (GPT-4V)~\cite{achiam2023gpt-4v} for additional pairwise comparison. For the two 3D assets, we render them from four or nine viewpoints. These two images will be concatenated together before passing into GPT-4V along with the text instructions. GPT-4V will return a decision of which of the two 3D assets is better according to the instruction. As shown in Fig.~\ref{fig:supplm_gpt4}, We evalute our results in three main criteria: text–asset alignment, 3D plausibility and texture details.

\section{Ethical Statement}
We confirm that all images used in this paper for research and publication have been obtained and used in a manner compliant with ethical standards. The individuals depicted in these images have given consent for their use, or the images are sourced from publicly available datasets and were used in accordance with the terms of use and permissions. Furthermore, the publication and use of these images do not pose any societal or ethical harm. We have taken necessary precautions to ensure that the research presented in this paper respects individual rights, including the right to privacy and the fundamental principles of ethical research conduct.

\begin{figure*}[!t]
    \centering
    \includegraphics[width=\linewidth]{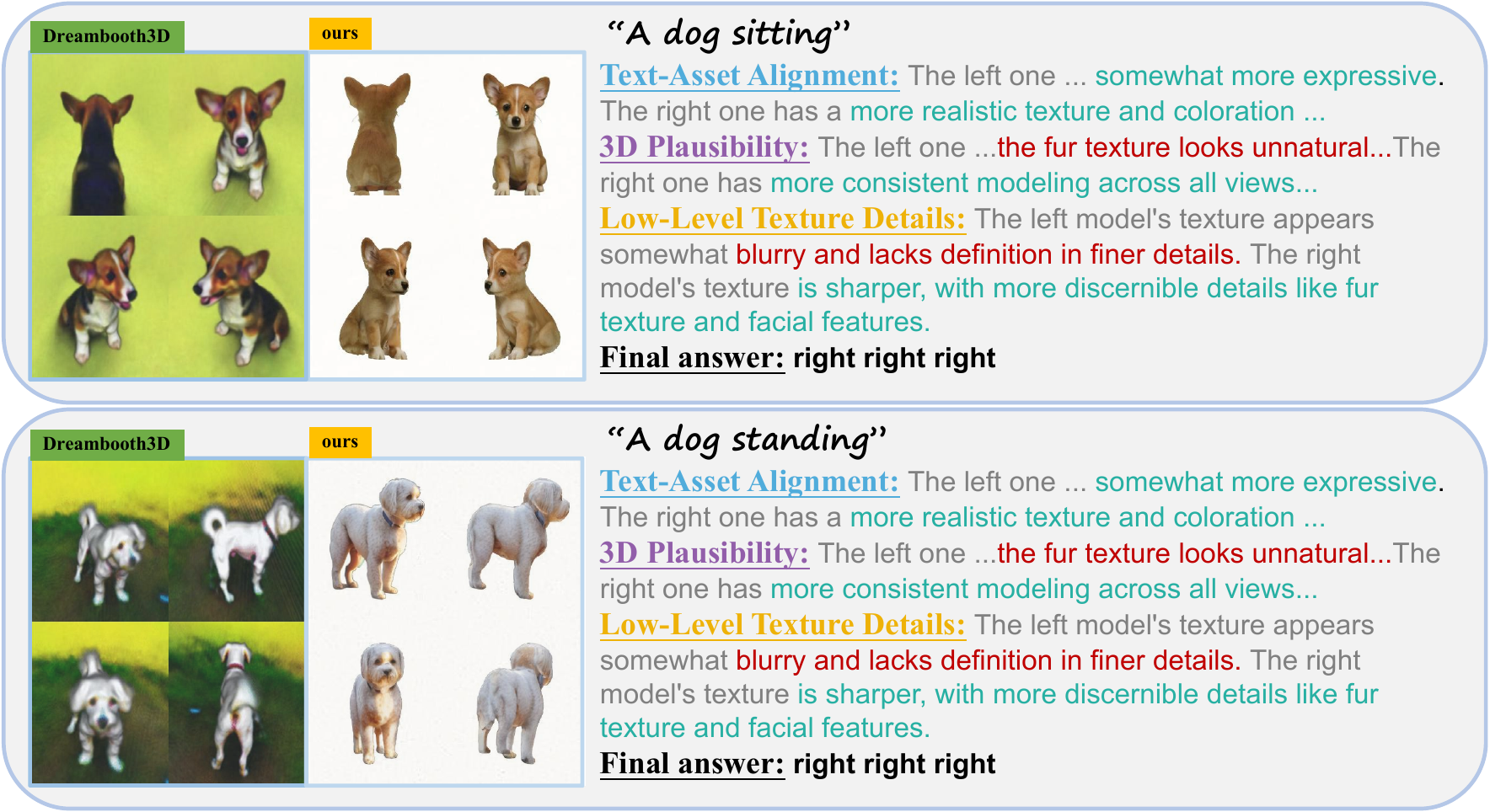}
    \vspace{-0.6cm}
    \caption{\textbf{More examples of the analysis by GPT-4V~\cite{achiam2023gpt-4v}.} }
    \label{fig:supplm_gpt4_1}
\end{figure*}

\end{document}